\pdfoutput=1

\documentclass[11pt]{article}

\usepackage[]{acl}

\usepackage{times}
\usepackage{latexsym}

\usepackage[T1]{fontenc}

\usepackage[utf8]{inputenc}

\usepackage{microtype}
\usepackage{amsfonts} 
\usepackage{tabularx}
\usepackage{multirow}
\usepackage{booktabs}
\usepackage{graphicx}
\usepackage{ragged2e}
\usepackage{footnote}
\usepackage{bbm}
\usepackage{bm}
\usepackage{latexsym}
\usepackage{amsmath}
\usepackage{amssymb}
\allowdisplaybreaks

\newcommand{\mixces}{\textsc{MixCE}}

\newcommand\bloomberg{$^\heartsuit$}
\newcommand\unc{$^\spadesuit$}
\newcommand\jhu{$^\clubsuit$}

\title{\mixces: Training Autoregressive Language Models \\ by Mixing Forward and Reverse Cross-Entropies}

\author{Shiyue Zhang\unc\thanks{\,\, Work done during an internship at Bloomberg.} $\;\;\;\;$ Shijie Wu\bloomberg $\;\;\;\;$ Ozan İrsoy\bloomberg \\ $\;\;\;\;$ \textbf{Steven Lu}\bloomberg $\;\;\;\;$ \textbf{Mohit Bansal}\unc $\;\;\;\;$ \textbf{Mark Dredze}\bloomberg\jhu $\;\;\;\;$ \textbf{David Rosenberg}\bloomberg \\
\bloomberg Bloomberg $\;\;$ \unc UNC Chapel Hill $\;\;$  \jhu Johns Hopkins University
}

\begin{document}
\maketitle

\begin{abstract}
Autoregressive language models are trained by minimizing the cross-entropy of the model distribution $Q_\theta$ relative to the data distribution $P$ -- that is, minimizing the \emph{forward cross-entropy}, which is equivalent to maximum likelihood estimation (MLE). We have observed that models trained in this way may ``over-generalize'', in the sense that they produce non-human-like text.  Moreover, we believe that \emph{reverse cross-entropy}, i.e., the cross-entropy of $P$ relative to $Q_\theta$, is a better reflection of how a human would evaluate text generated by a model. Hence, we propose learning with \mixces{}, an objective that mixes the forward and reverse cross-entropies. We evaluate models trained with this objective on synthetic data settings (where $P$ is known) and real data, and show that the resulting models yield better generated text \emph{without} complex decoding strategies.

\end{abstract}

\vspace{0.5em}
\hspace{.5em}\includegraphics[width=1.25em,height=1.25em]{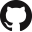}\hspace{.75em}\parbox{\dimexpr\linewidth-4\fboxsep-2\fboxrule}{\small{\url{https://github.com/bloomberg/mixce-acl2023}}}
\vspace{-.5em}

\section{Introduction}
\label{sec:intro}

Rapid advances in pre-trained large-scale autoregressive language models (LMs) have dramatically improved the performance of a variety of tasks~\cite{radford2019language, gpt3, opt, chowdhery2022palm}. 
However, these systems still struggle in many open-ended generation settings, where they are asked to produce a long text following a short prompt.
In these cases, we seek systems that generate sensical, coherent, fluent, and engaging, or in short, \emph{human-like} text~\cite{pillutla2022mauve}.

Different decoding strategies to generate such text from pretrained LMs suffer from different degeneration problems. Unbiased sampling\footnote{Unbiased sampling is vanilla random sampling, i.e., sampling with temperature=1.0. It is also called ancestral sampling~\cite{eikema-aziz-2020-map} or pure sampling~\cite{Holtzman2020The}. We call it unbiased sampling because it allows unbiased exploration of the model distribution.\label{footnote:sampling}} 
 usually results in incoherent and nonsensical text, while greedy and beam searches often get stuck in repetition loops~\cite{Holtzman2020The} (see examples in Figure~\ref{fig:problems}).
These observations suggest that the learned LM distribution $Q_\theta$ still differs substantially from the human LM distribution $P$.
A possible reason is that the autoregressive modeling of $Q_\theta$ gives a non-zero probability to every possible sequence of tokens, while many sequences are impossible under $P$. Nevertheless, we still hope that $Q_\theta(x)$ is as small as possible when $P(x)=0$. To this end, maximum likelihood estimation (MLE), i.e., minimizing the cross-entropy (CE) $-\mathbb{E}_{x \sim P}[\log Q_\theta(x)]$, is the most widely used objective to train $Q_\theta(x)$ using sequences sampled from $P$. 
In an idealized setting, with unlimited training data and model capacity, as well as a perfect optimizer, fitting $Q_\theta$ with MLE will learn a distribution as close to $P$ as we like.
However, in practice, we only have finite and noisy data. 

\begin{savenotes}
\begin{figure}[t]
    \centering
    \includegraphics[width=0.48\textwidth]{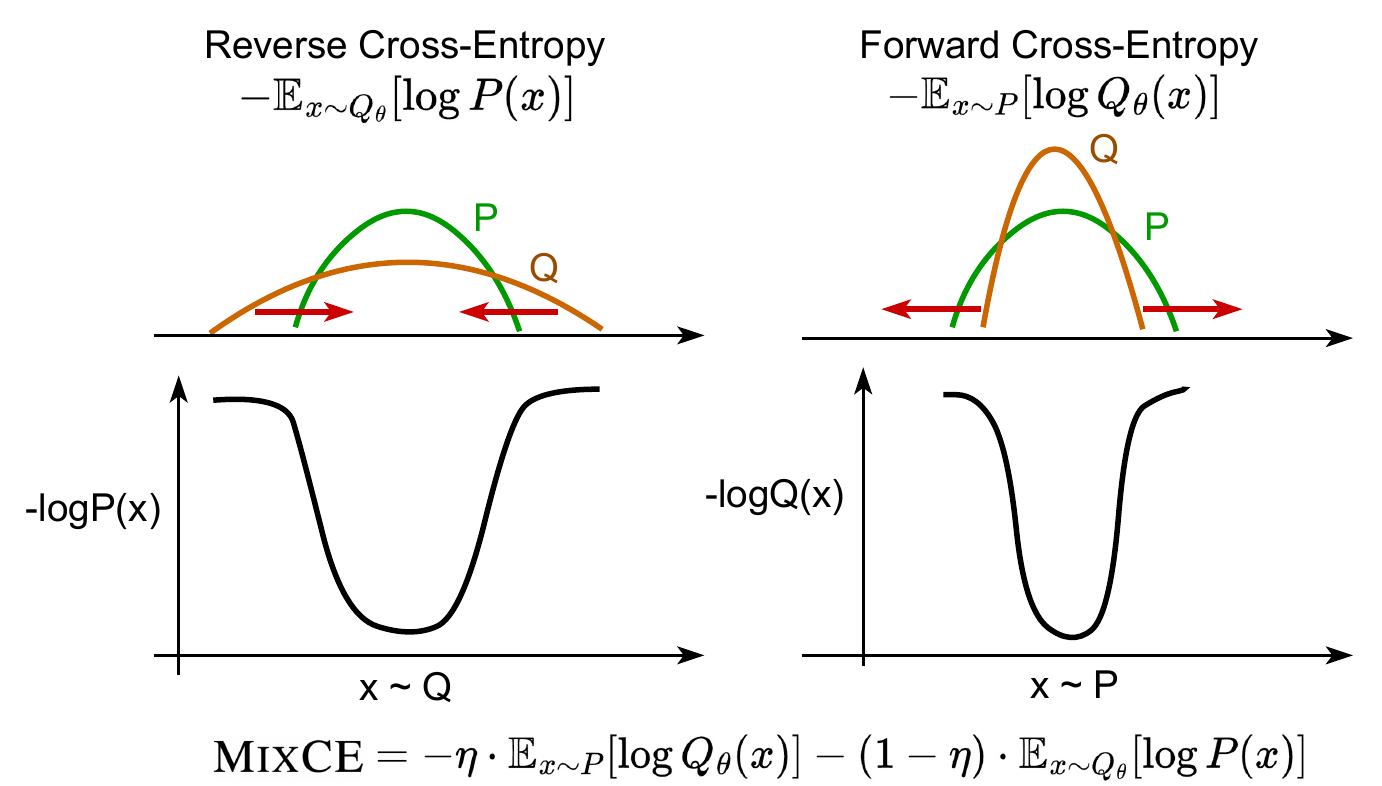}
    \vspace{-5pt}
    \caption{\mixces{} combines two complementary driving forces: reverse CE helps narrow the model distribution $Q_\theta$ down when it is broader than data distribution $P$, while forward CE helps broaden $Q_\theta$ out when it is narrower than $P$.\footnote{Note that $\log P(x)$ is infinite when $P(x)=0$. But in practice, we use $\log P(x) = \sum_t\log (P(x_t|x_{<t}) + \epsilon)$ to avoid $\log 0$ and $\epsilon=1e-30$.  }}
    \vspace{-7pt}
    \label{fig:mixces}
\end{figure}
\end{savenotes}

We argue that the MLE objective only weakly penalizes generations $x$ from $Q_\theta$ that are ``bad'', in the sense that $P(x)=0$. 
When $Q_\theta$ puts a small amount of probability mass onto $P(x)=0$ space, MLE cannot sufficiently discourage this behavior (see Figure~\ref{fig:mixces-app} in Appendix~\ref{app:for-rev}). Moreover, minimizing forward CE, $-\mathbb{E}_{x \sim P}[\log Q_\theta(x)]$, is equivalent to minimizing the forward KL divergence between $P$ and $Q_\theta$, i.e., KL$(P|| Q_\theta)$ = $\mathbb{E}_{x \sim P}[\log P(x)/Q_\theta(x)]$. Forward KL has a \emph{zero-avoiding} property -- avoiding $Q_\theta(x)=0$ when $P(x)\neq 0$~\cite{murphy2012machine}. Therefore, if there is noise in the data, $Q_\theta$ will try to cover the noise as well, 
which leads the model to \emph{over generalize}, in the sense of putting non-trivial probability mass over $P(x)=0$ generations~\cite{huszar2015not, theis2016note, pmlr-v80-ott18a, kang-hashimoto-2020-improved}. 
As a result, we observe samples from the model deviating from human-like text.
A common strategy is to modify the decoding method, e.g., top-$k$, top-$p$, typical, contrastive~\cite{fan-etal-2018-hierarchical, Holtzman2020The, meister+al.pre22, contrastive_decoding} samplings, to tailor the model distribution $Q_\theta$ in a post-hoc manner to avoid unwanted generations. 
In contrast, our approach differs: how can we obtain a better $Q_\theta$ to obviate the need for these sampling strategies? 

We propose a novel training objective for autoregressive LMs -- \mixces{} that \textbf{Mix}es the forward and reverse \textbf{C}ross-\textbf{E}ntropies: $- \eta \cdot \mathbb{E}_{x \sim P}[\log Q_\theta(x)] -(1 - \eta) \cdot \mathbb{E}_{x \sim Q_\theta}[\log P(x)]$. \mixces{} can be understood in two ways. First, we want model generations to be high-quality as well as diverse. Reverse cross-entropy reflects how we conduct human evaluations, sampling from the model $Q_\theta$ and evaluating it by the human $P$, where the focus is text \emph{quality}. Forward cross-entropy emphasizes the \emph{diversity} of model generations ~\cite{hashimoto-etal-2019-unifying}. Second, \mixces{} works similarly to a mixture of the forward and reverse KL divergences. The reverse KL divergence (KL$(Q_\theta || P)$) is \emph{zero-forcing} -- forcing $Q_\theta(x)=0$ when $P(x)=0$ -- and thus more strongly penalizes generating non-human-like samples compared to MLE. Overall, \mixces{} combines two complementary driving forces to better fit $Q_\theta$ to $P$ (Figure~\ref{fig:mixces}). We elaborate on these interpretations in \S~\ref{sec:math}.

Unfortunately, optimizing reverse cross-entropy is intractable because we do not know $P$. Hence, 
we propose an approximation of the reverse cross-entropy (see \S~\ref{sec:approximation}), which ends up being a \emph{self-reinforced} loss function that encourages the model to produce generations in which it is already confident. This loss function has the same computational complexity as forward cross-entropy, making \mixces{} easy to implement and as fast as MLE.

We demonstrate the effectiveness of \mixces{} in both a synthetic setting, where the ``human'' distribution $P$ is known, as well as a real setting. 
For the synthetic case, we evaluate six learning objectives: \mixces{}, \mixces{}$^*$ (\mixces{} without approximation), forward KL (=MLE), reverse KL, the mixture of two KL divergences, and Jensen–Shannon (JS) divergence. We show that \mixces{}$^*$ works slightly worse than the mixture of KLs while outperforming other objectives, and \mixces{} works worse than \mixces{}$^*$ but generally outperforms MLE. 
In real settings, we finetune GPT-2~\cite{radford2019language} of different sizes on three English text domains using \mixces{} or MLE. Our results show that, compared to MLE, unbiased sampling from \mixces{}-finetuned models produces text that has diversity~\cite{meister+al.pre22} closer to that of human text, has higher Coherence~\cite{su2022contrastive}, has higher Mauve~\cite{pillutla2021mauve}, and is preferred by humans. When using top-$p$ sampling~\cite{Holtzman2020The} and carefully tuning $p$, generations from MLE-finetuned models are similar to those generated from \mixces{}-finetuned models. Nonetheless, \mixces{} models have tuned $p$ values closer to 1, implying a less noisy model distribution. In addition, we modify the original Mauve to make it more robust to spurious features (e.g., text length), under which \mixces{} still improves over MLE when using unbiased sampling.

\section{Background and Related Work}
\subsection{Autoregressive Language Modeling}
Language generation is mostly based on the autoregressive language modeling methodology. 
The generation of one word is conditioned on previously generated words, $Q_\theta(x_t|x_{<t})$, and the final probability of the sequence $x$ is the product of probabilities of each step, $Q_\theta(x)=\prod_t Q_\theta(x_t|x_{<t})$. Early works build n-gram neural LMs~\cite{NIPS2000_728f206c} and then RNN-based LMs~\cite{mikolov2010recurrent}, and now Transformers~\cite{vaswani2017attention} have become the dominant architecture. 
Language generation models 
have either a decoder-only~\cite{mikolov2010recurrent} or an encoder-decoder architecture~\cite{sutskever2014sequence, DBLP:journals/corr/BahdanauCB14}. In this work, we focus 
on decoder-only LMs. In recent years, many large-scale pre-trained decoder-only LMs have been introduced~\cite{radford2019language, gpt3, opt, chowdhery2022palm}. They can be finetuned for downstream tasks and even perform surprisingly well in a zero-shot or few-shot manner. Despite the impressive performance, language \emph{degeneration} is one of the key issues that remain to be solved.

\subsection{Language \emph{De}generation}
According to \citet{Holtzman2020The}, language degeneration refers to output text that is \emph{bland, incoherent, or gets stuck in repetitive loops}. 
It is widely observed in open-ended generations 
from pretrained LMs. Please refer to some GPT2-large examples in Figure~\ref{fig:problems}. Two commonly observed patterns of degeneration are the incoherent text from unbiased sampling and the repetitive text from greedy or beam search.
Degeneration also appears in sequence-to-sequence generation tasks but in a slightly different form~\cite{stahlberg-byrne-2019-nmt}.

There is no agreement on what causes degeneration. \citet{pmlr-v80-ott18a} attribute it to data noise and the smooth class of model functions. 
It is inherent in the model's structure to have support everywhere, in particular, because all probabilities are produced by softmax, which is strictly positive. Therefore, \citet{hewitt2022truncation} assume that an LM distribution is the true data distribution plus a uniform-like smoothing distribution. 
Based on the observation that human-like text has a large but not too large likelihood under the learned LM distribution~\cite{zhang-etal-2021-trading}, a lot of works propose empirically useful decoding methods beyond unbiased sampling and greedy/beam search~\cite{fan-etal-2018-hierarchical, Holtzman2020The, eikema-aziz-2020-map, basu2021mirostat, meister+al.pre22, contrastive_decoding, hewitt2022truncation, su2022contrastive, rankgen22}. 
One of these approaches is the canonical top-$p$ (or nucleus) sampling method~\cite{Holtzman2020The}, which samples from top tokens that take up $p$ proportion (e.g., 95\%) of the probability mass at each decoding step.  Even though these decoding methods work impressively well, they are post-hoc fixes rather than learning the LM accurately in the first place. Therefore, some other works criticize the MLE training objective and propose alternative loss functions.

\subsection{Objectives Beyond MLE}
\label{sec:other_objective}

Unlikelihood training~\cite{Welleck2020Neural, li-etal-2020-dont} was proposed to penalize repetition (or any undesirable phenomenon) explicitly during training. The idea is to minimize the likelihood of a set of negative tokens at each generation step during training. The selection of negative tokens is pre-defined, e.g., tokens that appear often in the previous context. \mixces{} shares the same goal with unlikelihood training -- matching the human LM distribution, but provides a more general approach without targeting any specific problem. 

Similar to our motivation, \citet{kang-hashimoto-2020-improved} think that the zero-avoiding property of MLE makes the model sensitive to dataset noise. 
To cover these noisy examples, the model has to put non-trivial probability mass on the $P(x)=0$ area. To combat this problem, they propose a loss truncation method that drops high-loss (low-likelihood) examples during training time. 

\citet{Pang021} want to address the mismatch of learning objective and human evaluation (likelihood vs. quality) and introduce the GOLD algorithm to approximate reverse cross-entropy. Our approximation is similar to
theirs but has a different derivation process (see \S~\ref{sec:approximation}). Moreover, GOLD is evaluated on controlled generation tasks (e.g., summarization and translation) in which the goal is to generate one high-quality text for each input, and diversity is not so important. In contrast, if we train the LM only with reverse CE till convergence, the model will deterministically produce the most likely text for each prompt, which is undesirable for an LM. 
Therefore, mixing forward and reverse CEs is necessary.

The idea of \mixces{} is also relevant to GANs~\cite{goodfellow2014generative}. 
GANs optimize the Jensen–Shannon (JS) 
divergence between model and data distributions. Essentially, JS divergence is also for balancing the two driving forces of forward and reverse KL divergences~\cite{huszar2015not}, and it has been successfully used for evaluating LM-generated text~\cite{pillutla2021mauve}. 
However, probably due to the discrete nature of text, GANs have not been well applied to LM training. \citet{Caccia2020Language} show that previous language GANs often give up diversity for quality.

Another related work is \citet{popov2018fine}, which finetunes LMs with the sum of the forward cross-entropy loss and reverse KL divergence. They train a discriminator to estimate reverse KL, similar to a GAN. On the other hand, we directly approximate reverse cross-entropy in our objective function, without training an additional discriminator. 

Concurrently, with the same motivation as ours, \citet{ji2023tailoring} propose to replace MLE with minimization of the total variation distance (TVD) ~\cite{van2014probability} between data and model distributions. Notably, their final approximation of TVD, which they call TaiLr, is equivalent to forward cross-entropy when the hyperparameter $\gamma=0$ and equals our approximated reverse cross-entropy when $\gamma=1$.

\section{Methodology}
\subsection{\mixces{}}
\label{sec:math}
Our \mixces{} learning objective for training LMs is the combination of forward and reverse cross-entropies, written as 
\begin{equation}
\label{eq:oracle_mixces}
     -\eta \cdot \mathbb{E}_{x \sim P}[\log Q_\theta(x)] - (1 - \eta) \cdot \mathbb{E}_{x \sim Q_\theta}[\log P(x)]
\end{equation}
where $\eta$ is the mixing ratio. When $\eta=1$, it becomes the normal MLE objective; and when $\eta=0$, it is the reverse cross-entropy only. 

The \mixces{} loss can be understood in two ways. First, reverse and forward cross-entropy (CE) emphasize \emph{quality} and \emph{diversity} respectively. The reverse CE, $-\mathbb{E}_{x \sim Q_\theta}[\log P(x)]$, focuses on \emph{quality} because it resembles how we conduct human evaluations -- sampling from the model $Q_\theta$ and evaluating it by the human $P$. In human evaluations, the focus is more on the quality of the model-generated text. So, it is possible that a model always generates the same few high-quality texts, but still gets high human evaluation scores. This is similar to the \emph{mode collapse} problem of GANs. 
The forward CE, $-\mathbb{E}_{x \sim P}[\log Q_\theta(x)]$,  instead focuses more on \emph{diversity} because it needs any sample from $P$ to have a non-trivial probability under $Q_\theta$~\cite{hashimoto-etal-2019-unifying}. Note that it does not mean forward CE has zero effect on quality, rather, the model likelihood $Q_\theta(x)$ only loosely correlates with the human-perceived quality of $x$~\cite{zhang-etal-2021-trading}.

Second, we hypothesize that \mixces{} works similarly to a mixture of forward and reverse KL divergences, which we will show empirically in our synthetic experiments (\S~\ref{sec:synthetic}). On the one hand, minimizing forward KL is equivalent to optimizing forward CE. On the other hand, reverse KL divergence, $\mathbb{E}_{x \sim Q_\theta}[\log \frac{Q_\theta(x)}{P(x)}]$, has two parts: reverse CE and negative entropy of $Q_\theta$, $\mathbb{E}_{x \sim Q_\theta}[\log Q_\theta(x)]$. 
Reverse CE is minimized when the model deterministically outputs the most likely example, i.e., $Q_\theta(x)=\delta($the most likely $x$ under $P)$. Instead, minimizing the negative entropy (maximizing the entropy) of the model encourages it to be as uncertain as possible, i.e., having a large support and uniform distribution. This entropy term counteracts the narrowing-down effect of reverse CE. 
As discussed above, forward CE pushes the $Q$ distribution to fully cover the support of $P$. In this case, forward CE can also help counteract the narrowing-down effect of reverse CE, i.e., the maximizing entropy term becomes less important 
when forward CE is present. Hence, we think it is reasonable to drop it from reverse KL. 

Overall, \mixces{} combines two complementary training signals, as shown in Figure~\ref{fig:mixces}. Reverse CE prevents the model distribution from being broader than the data distribution, while forward CE is more helpful for preventing the model distribution from being narrower than the data distribution. Although forward CE also has non-zero loss when the model distribution is too wide, its loss magnitude is much smaller than what reverse CE provides (see Appendix~\ref{app:for-rev} for more discussion).  When data is clean, two CEs work jointly to help learn the data distribution better. When data is noisy, the mixing ratio $\eta$ allows us to trade-off between emphasizing a good coverage of the data and putting more weight on the actually high-quality sequences.

\subsection{Optimization of Reverse CE}
\label{sec:approximation}
Optimizing \mixces{} is non-trivial. 
The obstacle is to minimize the reverse CE, $-\mathbb{E}_{x \sim Q_\theta}[\log P(x)]$ with respect to $\theta$. To this end, we need to know $P$ and to have a differentiable sampling operation from $Q_\theta$.
In our synthetic experiments (\S~\ref{sec:synthetic}), we use a distribution $P$ of our own construction and use Gumbel-Softmax~\cite{JangGP17, MaddisonMT17} to make the sampling operation differentiable.

However, in a real setting, we do not know $P$. To deal with this, we take the following steps to derive an approximation of the gradient of the reverse cross-entropy (we omit the negative sign for simplicity):
\begin{align}
 &  \nabla_{\theta}\mathbb{E}_{x\sim Q_{\theta}}[\log P(x)] \label{eq:rev_ce}\\
 \approx  & \nabla_{\theta}\mathbb{E}_{x\sim Q_{\theta}}[P(x)] \label{eq:acc}\\
  = & \sum_{x} \nabla_{\theta}Q_{\theta}(x)P(x) \label{eq:policy1}\\
  = & \sum_{x}Q_{\theta}(x)\nabla_{\theta}\log Q_{\theta}(x)P(x) \label{eq:policy2}\\
   = & \sum_{x}P(x)Q_{\theta}(x)\nabla_{\theta}\log Q_{\theta}(x) \label{eq:importance1} \\
  = & \mathbb{E}_{x\sim P}[Q_{\theta}(x)\nabla_{\theta}\log Q_{\theta}(x)] \label{eq:importance2}\\
 = & \mathbb{E}_{x\sim P}[\prod_{t=1}^{T}Q_{\theta}(x_{t}|x_{<t})\sum_{t=1}^{T}\nabla_{\theta}\log Q_{\theta}(x_{t}|x_{<t})] \label{eq:unfold}\\
 \approx & \mathbb{E}_{x\sim P}[\sum_{t=1}^{T}Q_{\theta}(x_{t}|x_{<t})\nabla_{\theta}\log Q_{\theta}(x_{t}|x_{<t})] \label{eq:final}
\end{align}

First, from (\ref{eq:rev_ce}) to (\ref{eq:acc}), we substitute expected log-likelihood by \emph{expected accuracy}. \citet{ozan2019expected} shows that expected accuracy is a comparable or better alternative loss function to cross-entropy for classification tasks. 
Then, following the Policy Gradient theorem \cite{williams1992simple, sutton1999policy}, we get (\ref{eq:policy1}) and (\ref{eq:policy2}), where we view model $Q_\theta$ as the policy and $P(x)$ as the reward we want to optimize for the whole sequence. 
Next, we switch from the expectation of $Q_\theta$ to the expectation of $P$ (from (\ref{eq:policy2}) to (\ref{eq:importance1}) and (\ref{eq:importance2})), so that we can use the offline samples from $P$ (data samples in the training set) instead of online sampling from $Q_\theta$. 
We unfold $Q_\theta(x)$, which results in (\ref{eq:unfold}).
Until this point, theoretically, we are already able to optimize the model using Equation (\ref{eq:unfold}) without knowing $P$. However, the product of $Q_\theta(x_t|x_{<t})$ has a very high variance, and in practice, it underflows when $T$ is large. Therefore, we apply a final rough approximation that leads to (\ref{eq:final}). 

Equations (\ref{eq:unfold}) and (\ref{eq:final}) are apparently not equivalent to each other. Nonetheless, they have similar effects. Intuitively, in (\ref{eq:unfold}), we weigh the gradients of each sequence differently based on their sequence-level probabilities, $Q_\theta(x)$; in other words, it promotes high-likelihood sequences. Similarly, (\ref{eq:final}) weighs gradients at each step by $Q_\theta(x_t|x_{<t})$, i.e., promoting high-likelihood tokens at each step. So essentially, they both \emph{encourage the model to produce generations in which it is already confident}. We call it a \emph{self-reinforced} objective. 
To further illustrate why \emph{self-reinforcement} makes sense, we conduct an analysis using GPT-2~\cite{radford2019language}. Please refer to Appendix~\ref{app:self-reinforced} for a detailed discussion. In short, we show that MLE-pretrained GPT-2 on average assigns a higher probability to human text than to text sampled from the model. Therefore, when we promote high-probability sequences or tokens, it is like ``pushing'' the model distribution toward the human distribution. But, we need to avoid overly ``pushing'' it to the extremely high-probability region where repetitive greedy search outputs locate. 

Note that our approximation of reverse cross-entropy is relevant to the method proposed by~\citet{Pang021}, though we have a different derivation process from theirs. Please see Appendix~\ref{app:connection} for a detailed comparison.

Finally, combining forward CE and Equation (\ref{eq:final}), our approximated \mixces{} objective is to maximize  
\begin{equation}
    \mathbb{E}_{x\sim P}[\sum_{t=1}^{T}(\eta +(1-\eta) \cdot Q_{\theta}(x_{t}|\cdot))\nabla_{\theta}\log Q_{\theta}(x_{t}|\cdot)],
\end{equation}
where $Q_{\theta}(x_{t}|\cdot)$ is short for $Q_{\theta}(x_{t}|x_{<t})$. This loss function has the same computational complexity as forward CE (MLE). Since $Q_{\theta}(x_{t}|x_{<t})$ is strictly lower than 1 (it is around 0.017 to 0.13 when using GPT-2), the gradient from approximated reverse CE is smaller than that from forward CE. Therefore, it is important to tune $\eta$ to balance the effects of two CEs.

After the camera-ready version of this paper was published on ACL 2023~\cite{zhang-etal-2023-mixce}, we developed a new derivation of reverse CE that provides a more interpretable derivation of token-level self-reinforcement. Please refer to Appendix~\ref{app:alternative} for the details. In Appendix~\ref{app:ozan}, we also discuss the relation between \mixces{} and the leaky loss function proposed in \citet{ozan2019expected}. 

\section{Experiments}

\subsection{Synthetic Experiments}
\label{sec:synthetic}
We first conduct experiments in a synthetic ideal setting, where we know $P$, to show the effectiveness of mixing two cross-entropies with or without approximation. Moreover, during evaluation, we can directly compare the learned model parameters against the ground truth parameters of $P$. 

\paragraph{Define the ``human'' LM $P$.} We start by defining $P$ as a bi-gram LM. Bi-gram means that the prediction of the next token only depends on the immediately previous token, i.e., $P(x_t|x_{t-1})$. Therefore, $P$ is determined by a transition matrix among words $\mathbf{M} \in \mathbb{R}^{V \times V}$ ($V$=vocabulary size) and a start token probability distribution $\boldsymbol \pi \in \mathbb{R}^{V}$, i.e., stochastic finite-state automata. The last token in the vocabulary is the end-of-sequence (EOS) token. For simplicity, we initialize $\boldsymbol \pi$ as a uniform distribution. To initialize $\mathbf{M}$, we use two methods. The first is \textbf{random initialization}. We sample categorical
distributions from a Dirichlet ($\alpha$=0.5) prior to initialize each row of $\mathbf{M}$. However, one remaining problem 
is that $P$ has support everywhere. To have $P=0$ areas, we randomly assign 0s to a certain percent of values in each row of $\mathbf{M}$ and then re-normalize to sum to 1.\footnote{When we assign 0s, we make sure every token has non-zero transition probability to EOS.} We test 3 percentages: 10\%, 50\%, and 90\%. The second is \textbf{initialization using real data}. We sample 5000 pieces of text from WebText~\cite{radford2019language}, count the occurrence of bigrams, and then use the occurrence to initialize $\mathbf{M}$. 
In this case, there are naturally 0s in $\mathbf{M}$, and the larger the vocabulary size is, the sparser $\mathbf{M}$ is. No matter which initialization is used, we reserve the last row of $\mathbf{M}$ for EOS and it has all 0s, i.e., will not transit to any token. We set the vocabulary size $V$=20, 50, 100, 500, or 1000.\footnote{Our defined bi-gram LMs are always \emph{tight}, i.e., do not ``leak'' probability mass onto infinite sequences because we make sure that all accessible tokens also have non-zero paths to other tokens. Please refer to \citet{du2022measure} for the proof. \label{footnote:tight}}

\paragraph{Learn an LM $Q_\theta$.} We implement model $Q_\theta$ as a simple neural bigram LM. Given the word embedding $e_{i-1}$ of the previous token $x_{i-1}$, the next token is predicted via a simple neural network $f$:
$$h_{i-1} = \text{Dropout}(\text{ReLU}(\mathbf{W}_1e_{i-1} + \mathbf{b}_1)),$$
$$Q(x_i|x_{i-1}) = \text{Softmax}(\mathbf{W}_2h_{i-1} + \mathbf{b}_2),$$
where $\mathbf{W}_1 \in \mathbb{R}^{d \times d}$ ($d$ is the hidden dimension size), $\mathbf{b}_1 \in \mathbb{R}^d$, $\mathbf{W}_2\in \mathbb{R}^{d \times V}$, and $\mathbf{b}_2 \in \mathbb{R}^V$ are model parameters. After training this model, the learned transition matrix can be obtained by $\mathbf{M}'=f(\mathbf{E})$, $\mathbf{E}$ is the word embedding matrix. 

\paragraph{Synthetic data.} We sample sequences from $P$. We set the max sequence length as 500. 
We sample 50K and 5K sequences as the training and validation set, respectively. There is no test set because we directly compare the learned transition matrix $\mathbf{M}'$ to the gold $\mathbf{M}$ during evaluation.

\paragraph{Metrics.} (1) \textbf{avg. js}: we compute the JS divergence between each row (except the last row) of $\mathbf{M}'$ and the corresponding row in $\mathbf{M}$, and then average across rows. This metric evaluates the overall divergence of $\mathbf{M}'$ from $\mathbf{M}$, and equals 0 iff $\mathbf{M}'=\mathbf{M}$; (2) \textbf{avg. 0s}: we get the probabilities from $\mathbf{M}'$ from positions where the corresponding gold probabilities are 0 in $\mathbf{M}$, and take their average. If $\mathbf{M}'=\mathbf{M}$, avg. 0s = 0, but vice versa is not true. 

\begin{table}[t]
\centering
\small
\resizebox{0.48\textwidth}{!}{%
\begin{tabular}{ll|cc|cc}
\toprule
& & \multicolumn{2}{c}{Random (50\%)} & \multicolumn{2}{c}{WebText} \\
\cmidrule(lr){3-4} \cmidrule(lr){5-6} 
Vocab & Objective &  avg. js & avg. 0s  &  avg. js & avg. 0s \\
\midrule
& Gold & 0.0 & 0.0 & 0.0 & 0.0 \\
\midrule
20 & For. KL & 7.40e-4 & 1.44e-4 & 9.93e-4 & 1.79e-4\\ 
& Rev. KL & 1.36e-1 & \bf 7.42e-6 & 3.93e-3 & \bf 1.95e-6 \\
& Mix KLs & \bf 4.89e-4 & 5.15e-5 & \bf 9.91e-4 & 1.11e-5 \\
& JS & 2.14e-1& 4.88e-5 & 1.12e-2 & 5.84e-6 \\
& \mixces* & 8.12e-4& 1.05e-4 & 1.36e-3 & 1.19e-4\\
& \mixces & 7.02e-4 & 1.25e-4 & 1.00e-3 & 1.79-4\\
\midrule
50 & For. KL & 6.47e-3 & 5.65e-4  & 4.30e-3 & 4.77e-4\\ 
& Rev. KL & 4.29e-1& 1.53e-3 & 3.48e-2 & 5.30e-5\\
& Mix KLs & \bf 4.45e-3 & \bf 2.80e-4 & \bf 3.91e-3 & 2.83e-4 \\
& JS & 4.74e-1& 1.40e-3& 9.23e-3 & \bf 2.48e-5 \\
& \mixces* & 4.49e-3& 3.72e-4  & 3.94e-3 & 2.75e-4\\
& \mixces & 6.47e-3 & 5.64e-4 & 4.29e-3 & 4.77e-4 \\
\midrule
100 & For. KL & 3.56e-2 & 1.44e-3 & 9.70e-3 & 3.10e-4\\ 
& Rev. KL & 5.57e-1 & 3.62e-4  & 1.00e-1 & 4.04e-5\\
& Mix KLs & \bf 2.74e-2 & \bf 2.10e-4 & \bf 9.19e-3 & 1.84e-4\\
& JS & 5.53e-1 & 9.69e-4 & 1.73e-1 & 5.56e-4 \\
& \mixces* & 2.85e-2 & 9.16e-4 & 9.61e-3 & 1.87e-4\\
& \mixces & 3.56e-2 & 1.41e-3 & 9.69e-3 & \bf 3.16e-6\\
\midrule
500 & For. KL & 2.39e-1 & 1.49e-3 & 4.60e-2 & 1.78e-4\\ 
& Rev. KL & 6.78e-1 & \bf 2.76e-6  & 3.05e-1 & \bf 1.68e-5\\
& Mix KLs & \bf 2.32e-1  & 8.60e-4  & 4.27e-2 & 1.33e-4\\
& JS & 5.34e-1 & 7.19e-4 & 2.78e-1 & 3.84e-5\\
& \mixces* & 2.34e-1 & 1.38e-3 & \bf 4.23e-2 & 1.29e-4 \\
& \mixces & 2.35e-1 & 1.46e-3 & 4.53e-2 & 1.64e-4\\
\midrule
1000 & For. KL & 2.93e-1 & 8.80e-4 & 8.10e-2 & 1.50e-4\\ 
& Rev. KL & 6.85e-1 & \bf 1.21e-6 & 3.30e-1 & \bf 6.26e-6 \\
& Mix KLs & \bf 2.91e-1 & 8.57e-4 & 7.50e-2 & 1.17e-4\\
& JS & 4.59e-1 & 5.97e-4  & 3.02e-1 & 1.93e-5\\
& \mixces* & 2.92e-1 & 8.58e-4  & \bf 7.44e-2 & 1.14e-4\\
& \mixces & 2.92e-1& 8.76e-4 & 7.94e-2 & 1.42e-4 \\
\bottomrule
 \end{tabular}
 }
 \vspace{-5pt}
\caption{Synthetic experimental results. Random (50\%) randomly initializes $\mathbf{M}$ and sets 50\% of the probabilities to 0.  WebText means initializing $\mathbf{M}$ by the bigram occurrence in the WebText data. Gold refers to the results when $\mathbf{M}'$=$\mathbf{M}$. \emph{avg. js} is our main metric, which represents the average JS divergence between $\mathbf{M}$ and $\mathbf{M}'$ (please see the definition of \emph{avg. 0s} in text). Each number is a 5-seed average, and Table~\ref{table:ci} shows the 95\% confidence intervals of some experiments.} 
\label{table:syn}
\vspace{-5pt}
\end{table}

\begin{table*}[t]
\centering
\small
\resizebox{0.98\textwidth}{!}{%
\begin{tabular}{ll|cccc|cccc|cccc}
\toprule
& & \multicolumn{4}{c}{WikiText} & \multicolumn{4}{c}{WebText} & \multicolumn{4}{c}{WritingPrompts} \\
\cmidrule(lr){3-6} \cmidrule(lr){7-10} \cmidrule(lr){11-14} 
Model Size & Objective &  ppl & div & mauve & coh &  ppl & div & mauve & coh &  ppl & div & mauve & coh \\
\midrule
& Human & - & 0.89 & 1.0 & 0.628 & - & 0.84 & 1.0 & 0.633 & - & 0.85 & 1.0 & 0.473  \\
\midrule
\multirow{2}{*}{Small} & MLE & \bf 26.98 &  \bf 0.91 & 0.67 & 0.556 & \bf 21.45 & 0.87 & 0.90 & 0.555 & \bf 28.45 & 0.87 & 0.85 & 0.397 \\
& \mixces & 35.04 & \bf 0.87 & \bf 0.93 & \bf 0.567 & 21.69 & \bf 0.85 & \bf 0.92 & \bf 0.565 & 28.79 & \bf 0.86 & \bf 0.89 & \bf 0.403\\
\midrule
\multirow{2}{*}{Medium} & MLE & \bf 20.43 &  \bf 0.90 & 0.73 & 0.573 & \bf 15.92 & 0.87 & 0.88 & 0.560 & \bf 22.72 & 0.88 & 0.89 & 0.414 \\
& \mixces &  25.92 & \bf 0.88 & \bf 0.95 & \bf 0.584 & 16.51 & \bf 0.83 & \bf  0.93 & \bf 0.585 & 23.04 & \bf 0.86 & \bf 0.91 & \bf 0.419 \\
\midrule
\multirow{2}{*}{Large} & MLE &  \bf 18.24 &  \bf0.90 & 0.75 & 0.567 & \bf 14.13 & 0.87 & 0.81 & 0.570 & 21.95 & 0.87 & 0.87 & 0.425 \\
& \mixces & 23.44 & \bf 0.88 & \bf 0.95 & \bf 0.578 & 14.66 & \bf 0.82 & \bf 0.94 & \bf 0.592 & \bf 21.04 & \bf 0.86 & \bf 0.94 & \bf 0.429 \\
\bottomrule
 \end{tabular}
 }
 \vspace{-5pt}
\caption{Unbiased sampling results of models finetuned by MLE or \mixces{} on three datasets. For all metrics, the closer to the human scores the better. \textbf{Bold} numbers are the ones that are closer to human scores in each setting. Each number is a 3-run average.} 
\label{table:model-size}
\end{table*}

\begin{table*}[t]
\centering
\small
\resizebox{0.98\textwidth}{!}{%
\begin{tabular}{ll|cccc|cccc|cccc}
\toprule
& & \multicolumn{4}{c}{WikiText} & \multicolumn{4}{c}{WebText} & \multicolumn{4}{c}{WritingPrompts} \\
\cmidrule(lr){3-6} \cmidrule(lr){7-10} \cmidrule(lr){11-14} 
Model Size & Objective &  best $p$ & div & mauve & coh &  best $p$ & div & mauve & coh &  best $p$ & div & mauve & coh \\
\midrule
& Human & - & 0.89 & 1.0 & 0.628 & - & 0.84 & 1.0 & 0.633 & - & 0.85 & 1.0 & 0.473  \\
\midrule
\multirow{2}{*}{Small} & MLE & 0.85 & \bf0.89 & 0.93 & \bf 0.584 & 0.93 & \bf 0.84 & \bf 0.94 & \bf 0.580 & 0.97 & 0.86 & \bf 0.90 & \bf 0.410\\
& \mixces & \bf 0.99 & 0.87 & \bf 0.95 & 0.568 & \bf 0.99 & 0.84 & 0.93 & 0.571 & \bf 0.99 &  \bf 0.85 & \bf 0.90 & 0.407\\
\midrule
\multirow{2}{*}{Medium} & MLE &  0.85& \bf 0.88 & 0.95& \bf 0.602& 0.93 & \bf 0.85 & \bf 0.95 & 0.592 & 0.97 & 0.86 &\bf 0.92 & \bf 0.428\\
& \mixces & \bf 0.99 & 0.87 & \bf 0.96& 0.590 & \bf 0.99 & 0.81 & 0.93 & \bf 0.594 & \bf 0.99 &  \bf0.85 & \bf 0.92 & 0.427 \\
\midrule
\multirow{2}{*}{Large} & MLE & 0.87 & \bf 0.89 & 0.96 &\bf  0.594 & 0.95 & \bf 0.84 & 0.87 & 0.593 &  \bf 0.99 &  \bf 0.86 & 0.89 & 0.430\\
& \mixces &  \bf 0.99 & 0.87 &\bf  0.97 & 0.580  & \bf 0.99 & 0.81 & \bf 0.94 & \bf 0.601 &  \bf 0.99 &  \bf 0.86 &  \bf 0.94 &  \bf 0.435\\
\bottomrule
 \end{tabular}
 }
 \vspace{-5pt}
\caption{Top-$p$ sampling results of the same models as Table~\ref{table:model-size}. Since changing the decoding method will not affect perplexity, we report the selected best $p$ instead.  } 
\label{table:topp}
\vspace{-5pt}
\end{table*}

\paragraph{Objectives.} (1) \textbf{Forward KL}, KL$(P|| Q_\theta)$ = $\mathbb{E}_{x \sim P}[\log P(x)/Q_\theta(x)]$, which is equivalent to MLE; (2) \textbf{Reverse KL}, KL$(Q_\theta|| P)$ = $\mathbb{E}_{x \sim Q_\theta(x)}[\log Q_\theta(x)/P(x)]$; (3) \textbf{Mixture of two KLs}, $\eta \ \cdot$ KL$(P|| Q_\theta$) + (1 - $\eta$) $\cdot$ KL$(Q_\theta|| P)$; (4) \textbf{JS}, we use a general definition of JS divergence~\cite{huszar2015not}, $\eta\ \cdot$ KL$(P|| M$) + (1 - $\eta$) $\cdot$ KL$(Q_\theta|| M)$, where $M$=$\eta \ \cdot$ P + (1 - $\eta$) $\cdot$ $Q_\theta$;\footnote{When $\eta=0.5$, it is the same as the objective of GAN~\cite{goodfellow2014generative}. But instead of using GAN's min-max loss, we directly optimize JS because we know $P$.} (5) \textbf{Oracle mixture of cross-entropies} (\mixces{}$^*$), where we use the known $P$. (6) \textbf{Approximated mixture of cross-entropies} (\mixces{}), where we assume $P$ is unknown. Except for Forward KL and \mixces{}, the other four objectives all need to sample from $Q_\theta$ and require gradients to pass through this sampling operation. To this end, we use Gumbel-Softmax~\cite{JangGP17, MaddisonMT17} to make sampling differentiable. 

\paragraph{Model selection.} During training, we check the validation loss (the value of the objective function) after every epoch and only save the best checkpoint that has the lowest validation loss.
For objectives with $\eta$, we choose the best $\eta$ based on the avg. js result on the validation set. We report a 5-seed average for each experiment. The search space of $\eta$ is [0.99, 0.9, 0.5, 0.1, 0.01]. Selected best $\eta$s are reported in Table~\ref{table:syn-eta} in the Appendix.

\paragraph{Results.} Table~\ref{table:syn} (and  Table~\ref{table:syn-app} in the Appendix) shows the results of our synthetic experiments. Across 4 kinds of initialization of $\mathbf{M}$ and 5 vocabulary sizes, we observe some common patterns. First, the mixture of two KLs often gets the best avg. js compared to other objectives, and \mixces{}$^*$ usually comes second. This supports our expectation that the mixture of two cross-entropies approximates the mixture of two KLs (\S~\ref{sec:math}), as well as demonstrates that combining two KLs or CEs can help learn the data distribution more accurately compared to MLE. Second, the approximated \mixces{} usually under-performs \mixces{}$^*$ but outperforms forward KL (MLE). Third, reverse KL generally works best for the avg. 0s metric, due to its property of \emph{zero-forcing} -- forcing $Q_\theta(x)=0$ when $P(x)=0$. Lastly, JS divergence oftentimes works similarly to reverse KL, which is consistent with the observation made by ~\citet{Caccia2020Language} -- language GANs trade off diversity for quality.

\subsection{GPT-2 Experiments}
Next, we test \mixces{} in a real setting where we do not know $P$, but we have finite samples from $P$. We use GPT-2~\cite{radford2019language} as the LM $Q_\theta$. Though GPT-2 models are already pre-trained by MLE, for simplicity, we use different objectives to finetune it. We test GPT-2 in 3 sizes: small (24M), medium (355M), and large (774M). See more implementation details in Appendix~\ref{app:repro}.

\paragraph{Real data.} We use English text data from 3 domains: (1) WikiText~\cite{merity2016pointer}: text from Wikipedia; (2) WebText~\cite{radford2019language}: text from the Web. It was used for pretraining GPT-2; and (3) WritingPrompts~\cite{fan-etal-2018-hierarchical}: text from the writing prompts forum of Reddit. We sample from each of these 3 datasets to form our training, development, and test sets. By default, our training/development/test set contains 50K/5K/5K examples.  
Please find more details about these datasets in Appendix~\ref{app:repro}.

\paragraph{Metrics.} (1) \textbf{Perplexity (ppl)} is defined as $e^{-\frac{1}{N*T}\sum_N \sum_T log_eQ_\theta(x_t|x_{<t})}$, where $N$ is the number of examples and $T$ is the sequence length. Perplexity is not necessarily correlated with human perceived quality~\cite{zhang-etal-2021-trading}.  (2)
\textbf{Diversity (div)}: following~\citet{meister+al.pre22}, we define $n$-gram diversity as the average fraction of unique vs. total n-grams for $n \in$ \{1, 2, 3, 4\} in each piece of text. (3) \textbf{Mauve}~\cite{pillutla2021mauve} compares model-generated text against human text via a KL divergence curve and is the state-of-the-art metric for open-ended text generation. We use Mauve as our primary metric. (4) \textbf{Coherence (coh)}~\cite{su2022contrastive} computes the cosine similarity between the embedding of prompt and the embedding of continuation, and embeddings are from SimCSE~\cite{gao-etal-2021-simcse}. All metrics are \emph{the closer to human scores the better}.

\paragraph{Objectives.} Since we have no access to $P$, we can only implement two out of the six objectives we test in the synthetic setting: (1) \textbf{MLE}, which is equal to forward CE or forward KL; (2) \textbf{\mixces{}}, the approximated mixture of cross-entropies. 

\paragraph{Decoding.} We use \textbf{unbiased sampling} (see footnote~\ref{footnote:sampling}) as our primary decoding method as it allows us to explore
the learned distribution in an unbiased way~\cite{eikema-aziz-2020-map}. Additionally, we test \textbf{top-$p$ sampling}~\cite{Holtzman2020The} to check if \mixces{} is complementary to advanced decoding methods, and we carefully tune $p$ on the development set. For each text, we take the first 50 tokens (by GPT-2 tokenizer) as the prompt and set the  max generation length as 512.

\paragraph{Model selection.} We finetune the model for 5 epochs on the training set and save the best checkpoint with the lowest dev loss. We select the best mixing ratio $\eta$ and the best $p$ based on the Mauve score on the dev set. The search space of $\eta$ is [0.99, 0.9, 0.7, 0.5, 0.3, 0.1, 0.01, 0.0] and that of $p$ is [0.85, 0.87, 0.89, 0.91, 0.93, 0.95, 0.97, 0.99].  Selected best $\eta$s are reported in Table~\ref{table:gpt2-eta} in the Appendix. Best $p$s are reported in Table~\ref{table:topp}. Metric scores are reported on the test set and are 3-run averages because sampling is stochastic.

\paragraph{Results.} Table~\ref{table:model-size} shows unbiased sampling results of models in different sizes and finetuned with different objectives on three datasets. As you can see, \mixces{}-finetuned models usually get worse perplexity but consistently better diversity, mauve, and coherence, compared to MLE-finetuned models. Table~\ref{table:topp} shows top-$p$ sampling results from the same models as Table~\ref{table:model-size}. Since perplexity will not change as the decoding method changes, we instead report the selected best $p$ in this table. 
It can be seen that after carefully applying top-$p$ sampling, \mixces{}-finetuned models work on par with MLE-finetuned models for diversity, mauve, and coherence. Nonetheless, the best $p$ for \mixces{} models is always 0.99, while MLE models have smaller and more diverse $p$s. This indicates that \mixces{} leads to a less noisy model distribution.  
\begin{savenotes}
\begin{table}[t]
\centering
\small
\begin{tabular}{l|ccc}
\toprule
& \multicolumn{3}{c}{Which is better?} \\
\cmidrule(lr){2-4}  
Dataset & \mixces{} & MLE & Same \\
\midrule
WikiText & 135* & 85 & 95 \\
WebText & 139* & 79 & 97 \\
WritingPrompts & 111 & 119 & 85\\
\bottomrule
 \end{tabular}
 \vspace{-5pt}
\caption{Human evaluation results. The star (*) means significantly\footnote{The significance test is conducted following the bootstrap test setup \cite{efron1994introduction}.} better ($p<$0.01).} 
\label{table:human_eval}
\vspace{-5pt}
\end{table}
\end{savenotes}
 
\paragraph{Human evaluation.} Besides automatic metrics, we also conduct a human evaluation. Following~\citet{rankgen22}, we conduct blind A/B testing. We randomly sample 105 examples from each dataset. For each example, we ask humans to read two generations from MLE and \mixces{}-finetuned GPT-2 large models, respectively, and the order of showing these two generations is random. 
We use unbiased sampling to get the generations. 
Then, we ask them to judge which one is better (or they are the same) and justify their preference, based on fluency, coherence, informativeness, and whether it is sensical. We conduct this evaluation on Amazon Mechanical Turk and collect 3 responses for each example. Please refer to Appendix~\ref{app:human_eval} for more details and examples. The final results are shown in Table~\ref{table:human_eval}. As you can observe, \mixces{}-finetuned models significantly outperform MLE-finetuned models on both WikiText and WebText domains, while the two methods perform similarly on WritingPrompts. It is also worth noting that, compared to the results shown in Table~\ref{table:model-size}, none of the 4 automatic metrics share the same trend with human evaluation.

\subsection{Robustness \& Analysis}

\paragraph{Varying training data sizes.} We test 3 other training data sizes: 10K, 25K, and 100K using GPT-2 small. Table~\ref{table:data-size} in the Appendix contains the results, and it shares the same story trend as Table~\ref{table:model-size}: \mixces{}-finetuned models get worse perplexity but in general work better than MLE-finetuned models for diversity, mauve, and coherence. \vspace{2pt}

\paragraph{Varying $\eta$ and max generation length.} To examine how the mixing ratio $\eta$ and the max generation length affect the performance, we show the mauve score curves on the dev set in Figure~\ref{fig:res_dev}. The x-axis is the mixing ratio $\eta$ from 0 to 1 (\mixces{}=MLE when $\eta=1$), and the y-axis is the mauve score with different max generation lengths (128, 320, and 512). First, reasonable performances are usually observed when $\eta \ge 0.1$, and only training the models with approximated reverse CE (i.e., $\eta=0$) leads to degeneration. Second, the advantage of \mixces{} is more prominent when the max generation length is longer. 

\paragraph{Controlled Mauve.} The max generation length is not the actual text length because when sampling from the model, EOS can be generated at any step. We find that the actual \emph{text length} can affect the mauve computation. Even if we truncate all texts to the same length, the \emph{incompleteness} caused by truncation can be another confounding factor. Both text length and text completeness are irrelevant to text quality but can be used by mauve to distinguish model generations from human texts.
Therefore, to eliminate the influence of these confounding factors, we propose a \emph{controlled mauve} (or \emph{c-mauve}) computation approach. Concretely, for human texts and model generations, we randomly sample 10K $L$-length text fragments from each of these two sets. $L$ is the number of tokens. Then, we compute the mauve between these two sets of fragments. 
Table~\ref{table:cotrolled-mavue} shows the results. 
As you can see, c-mauve scores are in general very high ($\geq 0.90$), which may indicate that, after controlling the confounding factors, the ability of mauve to distinguish model text from human text has been weakened. \mixces{} still gets better performance than MLE in most cases. Besides, we also compute controlled coherence in the same fashion, and \mixces{} retains its advantage. Please refer to Appendix~\ref{app:controlled-coh} for more details about controlled Mauve and Coherence.

\section{Conclusion}
We propose a novel training objective, \mixces{}, for autoregressive language modeling. \mixces{} combines forward and reverse cross-entropies, which can be viewed as combining two complementary driving forces for better fitting the model distribution to the data distribution. We demonstrate the  superiority of \mixces{} over MLE in both synthetic and real settings via both automatic and human evaluations. In the future, \mixces{} can be potentially used for pretraining language models.

\section*{Acknowledgments}
We thank anonymous reviewers for their valuable comments. We thank Xiang Zhou for the helpful discussions. This work was supported by a Bloomberg Data Science Ph.D. Fellowship.

\section*{Limitations}
One apparent disadvantage of \mixces{} is the mixing ratio $\eta$. As shown in Table~\ref{table:gpt2-eta} and Figure~\ref{fig:res_dev}, the best $\eta$ changes as the experimental setting changes. It may be because we use mauve as the model selection criteria or because different datasets have different noise levels. In general, we do not have a good answer to which $\eta$ should be used. The ideal solution is to select $\eta$ based on the performance of the development set like what we did. However, in pretraining settings, it is too expensive to search over multiple $\eta$s. Therefore, how to find a universal $\eta$ or how to determine $\eta$ automatically is an important problem to resolve before \mixces{} can be reliably used for pretraining. 

As we mentioned in \S~\ref{sec:intro}, language degeneration of open-ended generation shows two distinct patterns: the nonsensical text from unbiased sampling and the repetition loops from greedy search. Though \mixces{} helps improve the performance of sampling, we still see repetition loops when using greedy search.

\section*{Ethical Considerations}
As the OpenAI team pointed out, GPT-2 does not distinguish fact from fiction, so it can not support use cases that require the generated text to be true.
Additionally, GPT-2 reflect the biases inherent to the systems they were trained on, so it can not be deployed into systems that interact with humans unless the deployers first carry out a study of biases relevant to the intended use case. Though our \mixces{}-finetuned GPT-2 gets improved performance with respect to the metrics we used, the above statement still holds. At this point, we are not sure whether \mixces{} can help improve factuality or lead to less biased generations, but we are sure that the generations still have non-factual content and biases.  

\bibliography{anthology,custom}
\bibliographystyle{acl_natbib}

\appendix
\section*{Appendix}

\begin{figure*}[t]
    \centering
    \includegraphics[width=0.90\textwidth]{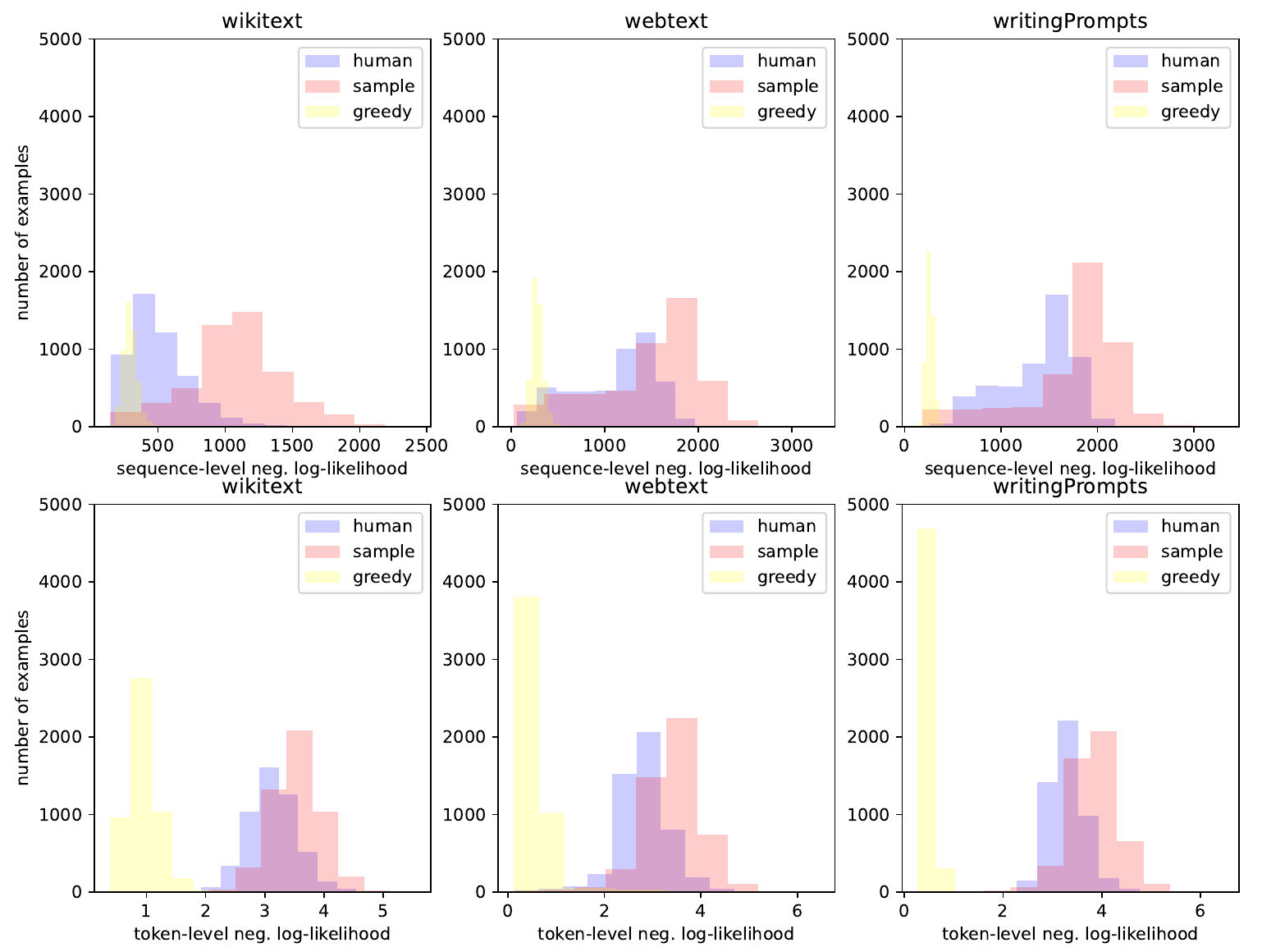}
    \caption{The histograms of sequence-level and token-level negative log-likelihoods of human texts and model generations from GPT-2 large. }
    \label{fig:neg_log_likelihood}
\end{figure*}

\begin{table*}
\centering
\small
\resizebox{0.9\textwidth}{!}{%
\begin{tabular}{ll|cccc|cccc|cccc}
\toprule
& & \multicolumn{4}{c}{WikiText} & \multicolumn{4}{c}{WebText} & \multicolumn{4}{c}{WritingPrompts} \\
\cmidrule(lr){3-6} \cmidrule(lr){7-10} \cmidrule(lr){11-14} 
Data Size & Objective & ppl & div & mauve & coh &  ppl & div & mauve & coh &  ppl & div & mauve & coh \\
\midrule
& Human & - & 0.89 & 1.0 & 0.628 & - & 0.84 & 1.0 & 0.633 & - & 0.85 & 1.0 & 0.473  \\
\midrule
\multirow{2}{*}{10K} & MLE & \bf 29.23 & 0.91 & 0.60 & 0.537 & \bf 22.03 & 0.88 & 0.82 & 0.542 & \bf 30.40 & 0.88 & 0.74 & 0.385 \\
& \mixces &  36.70 & \bf 0.88 & \bf 0.93 & \bf 0.546 & 22.79 & \bf 0.83 & \bf 0.86 & \bf 0.562 & 30.65 & \bf 0.87 & \bf 0.81 & \bf 0.395 \\
\midrule
\multirow{2}{*}{25K} & MLE &  \bf 27.90 & 0.91 & 0.68 & 0.545 & \bf 21.75 & 0.88 & 0.86 & 0.547 & \bf 29.37 & 0.88 & 0.79 & 0.394 \\
& \mixces & 35.73 & \bf 0.88 & \bf 0.94 & \bf 0.562 & 21.97 & \bf 0.85 & \bf 0.88 & \bf 0.561 & 29.67 & \bf 0.86 & \bf 0.86 & \bf 0.401 \\
\midrule
\multirow{2}{*}{100K} & MLE &  \bf 25.93 & \bf 0.90 & 0.69 & 0.559 & \bf 21.31 & 0.87 & 0.90 & 0.556 &\bf  27.63 & 0.87 & 0.88 & 0.401 \\
& \mixces &  34.13 & 0.87 & \bf 0.93 & \bf 0.575 & 21.58 & \bf 0.85 & \bf 0.92 & \bf 0.566 & 28.01 & \bf 0.85 & \bf 0.90 & \bf 0.409 \\
\bottomrule
 \end{tabular}
 }
\caption{Unbiased sampling results of GPT-2 small models finetuned by MLE or \mixces{} on three datasets of different training data sizes. All metrics are the closer to the human scores the better. \textbf{Bold} numbers are the ones that are closer to human scores in each setting. } 
\label{table:data-size}
\end{table*}

\section{Connection to \citet{Pang021}}
\label{app:connection}
In Section~\ref{sec:approximation}, we introduce an approximation of the reverse cross-entropy (CE) objective. Similarly, \citet{Pang021} also propose to approximate reverse CE, and the resulting GOLD algorithm is similar to our Equation~\ref{eq:final}. Here, we would like to clarify the difference and connection.

The following equation is the start policy gradient equation used by \citet{Pang021}.
$$\mathbb{E}_{\tau \sim \pi_\theta}[\sum_t\nabla_\theta \log \pi_\theta(a_t|s_t)\hat{Q}(s_t, a_t)]$$
They used different notations from ours. $\pi_\theta$ is the same as our $Q_\theta$, i.e., $\pi_\theta(a_t|s_t)$ is the same as our $Q_\theta(x_t|x_{<t})$. $\hat{Q}$ is the accumulated future reward from timestamp $t$, $\sum_{t'=t}^T\gamma^{t'-t}r_{t'}$, $\gamma$ is the decay factor and $r_{t'}$ is the reward for each step.  We will discuss $\hat{Q}$ in detail later.

Then, they apply importance sampling to sample from a different behavioral policy $\pi_b$. Since they also use examples from the training set, their $\pi_b$ is the same as our human (or data) distribution $P$. 
$$\mathbb{E}_{\tau \sim \pi_b}[\sum_t w_t \nabla_\theta \log \pi_\theta(a_t|s_t)\hat{Q}(s_t, a_t)]$$
$w_t$ is the importance weight. They use a per-action approximation: $w_t \approx \frac{\pi_\theta(a_t|s_t)}{\pi_b(a_t|s_t)}$, which is similar to how we get Equation~\ref{eq:final} from Equation~\ref{eq:unfold}.

Since $\pi_b$ is unknown, they assume a uniform distribution: $\pi_b \approx 1/N$ ($N$ is the number of training examples). Hence, their final approximated gradient is:
$$\mathbb{E}_{\tau \sim \pi_b}[\sum_t \pi_\theta(a_t|s_t) \nabla_\theta \log \pi_\theta(a_t|s_t)\hat{Q}(s_t, a_t)]$$
They define $r_{t'}$ and $\hat{Q}$ in three ways. The first is called $\delta$-reward, i.e., $\hat{Q}=1$. In this case, their final gradient is exactly the same as our Equation~\ref{eq:final}. However, as you can see, we take a different path of derivation. Instead of using this $\delta$-reward, our $\hat{Q}$ is the sequence-level reward $P(x)$. The reward $P(x)$ nicely helps us to switch from the expectation of $Q_\theta$ to the expectation of $P$ (from Equation~\ref{eq:policy2} to Equation~\ref{eq:importance2}). Therefore, without assuming a uniform distribution of $\pi_b$, our $\pi_b$ is just $P$. 

When using the other two rewards, they also need to know $P$. To address this, they use an MLE-pretrained model as a proxy of $P$. 

Overall, we introduce a different derivation approach for approximating reverse CE. Moreover, as we mentioned in \S~\ref{sec:other_objective}, \citet{Pang021} focused on improving controlled generation tasks where the focus is on the quality of the text, while we focus on open-ended generations where quality and diversity are both important. Therefore, we mix reverse CE with forward CE to form our \mixces{} learning objective.

\section{Alternative Derivation of Reverse CE}
\label{app:alternative}
Here we propose another derivation of approximate reverse cross-entropy (CE) (Eq.~\ref{eq:rev_ce}, which also results in Eq.~\ref{eq:final}). Different from the derivation in Section~\ref{sec:approximation}, here we provide a more intuitive interpretation of how we end up with token-level self-reinforcement (Eq.~\ref{eq:final}).

\begin{align}
& \mathbb{E}_{x\sim Q_{\theta}}[\log P(x)] \\
 = & \mathbb{E}_{x\sim Q_{\theta}}\left[\sum_{t=1}^{T}\log P(x_{t}\mid x_{<t})\right]\\
 = & \sum_{t=1}^{T}\mathbb{E}_{x\sim Q_{\theta}}\left[\log P(x_{t}\mid x_{<t})\right]\\
 = & \sum_{t=1}^{T}\mathbb{E}_{x_{<t}\sim Q_{\theta}}\mathbb{E}_{x_{t}\sim Q_{\theta}(\cdot\mid x_{<t})}\left[\log P(x_{t}\mid x_{<t})\right]\\
 \approx & \sum_{t=1}^{T}\mathbb{E}_{x_{<t}\sim Q_{\theta}}\mathbb{E}_{x_{t}\sim Q_{\theta}(\cdot\mid x_{<t})}\left[P(x_{t}\mid x_{<t})\right] \label{eq:alter_approx1}\\
 \approx  & \sum_{t=1}^{T}\mathbb{E}_{x_{<t}\sim P}\mathbb{E}_{x_{t}\sim Q_{\theta}(x_{t}\mid x_{<t})}P(x_{t}\mid x_{<t}) \label{eq:alter_approx2}
\end{align}

Then, we calculate the gradient as follows:
\begin{align}
& \sum_{t=1}^{T}\mathbb{E}_{x_{<t}\sim P}\nabla_{\theta}\mathbb{E}_{x_{t}\sim Q_{\theta}(x_{t}\mid x_{<t})}P(x_{t}\mid x_{<t}) \\
= & \sum_{t=1}^{T}\mathbb{E}_{x_{<t}\sim P}\mathbb{E}_{x_{t}\sim P(x_{t}\mid x_{<t})}[Q_{\theta}(\cdot)\nabla_{\theta}\log Q_{\theta}(\cdot)]\footnotemark  \\
= & \sum_{t=1}^{T}\mathbb{E}_{x\sim P}[Q_{\theta}(x_{t}\mid x_{<t})\nabla_{\theta}\log Q_{\theta}(x_{t}\mid x_{<t})]\\
= & \mathbb{E}_{x\sim P}[\sum_{t=1}^{T}Q_{\theta}(x_{t}\mid x_{<t})\nabla_{\theta}\log Q_{\theta}(x_{t}\mid x_{<t})] \label{eq:final2}
\end{align}
\footnotetext{$Q_{\theta}(\cdot) = Q_{\theta}(x_{t}\mid x_{<t})$}

In this derivation, we make two approximation steps. Following the same idea as Eq.~\ref{eq:acc}, we get Eq.~\ref{eq:alter_approx1} by substituting expected log-likelihood with expected accuracy~\cite{ozan2019expected}; but differently, here, they are log-likelihood or accuracy of tokens at step $t$. Then, we propose to approximate the model prefix ($x_{<t}\sim Q_{\theta}$) by human prefix ($x_{<t}\sim P$), which results in Eq.~\ref{eq:alter_approx2}. The final Eq.~\ref{eq:final2} is exactly the same as Eq.~\ref{eq:final}. It is also worth noting that, compared to the derivation in Section~\ref{sec:approximation}, in this alternative derivation, we approximate the loss function instead of the gradient, which is more intuitive in the sense that we know what we are exactly optimizing.  


\section{Relation to \citet{ozan2019expected}}
\label{app:ozan}
In our derivation described in Section~\ref{sec:approximation}, we approximate Eq.~\ref{eq:rev_ce} by Eq.~\ref{eq:acc}. However, after this step, it is no longer reverse CE anymore; instead, it is optimizing the expected accuracy of model $Q_\theta$ because $\mathbb{E}_{x\sim Q_{\theta}}[P(x)]=\mathbb{E}_{x\sim P}[Q_{\theta}(x)]$. Optimizing the expected accuracy is the exact proposal of \citet{ozan2019expected} for conducting classification tasks. Quoting their original words, they think the benefit of \emph{expected accuracy} over \emph{expected log-likelihood} is ``instead of prioritizing correction of those instances that we perform very poorly on, by weighing probability errors equally, we might just be able to push more instances to the other side of the decision boundary.'' This is the same motivation of our work: we hope the reverse CE loss (with or without approximation) can focus on ``quality'' rather than ``diversity''. Meanwhile, since \emph{expected accuracy} has close-to-zero derivatives for examples that we are the most incorrect on, \citet{ozan2019expected} also proposes to mix expected accuracy with expected log-likelihood, which is similar to our \mixces{} objective. 

But different from their work, we focus on text generation instead of classification. We need another critical approximation step (either Eq.~\ref{eq:unfold} to Eq.~\ref{eq:final} in Section~\ref{sec:approximation} or Eq.~\ref{eq:alter_approx2} in Appendix~\ref{app:alternative}) to arrive at the final token-level self-reinforcement loss function.

\section{Intuition behind the Self-reinforced Objective}
\label{app:self-reinforced}

To further illustrate why this \emph{self-reinforced} objective (Equation (\ref{eq:unfold}) or (\ref{eq:final})) makes sense and their shortcomings, we conduct an analysis using GPT-2 large~\cite{radford2019language}. We first sample 5000 pieces of text from WikiText, WebText, and WritingPrompts, respectively, and we call them \emph{human} texts.
Then, using the first 50 tokens of each human text as a prompt, we get 5000 sampling and greedy search generations from pretrained GPT-2 large (max generation length = 512). Next, we use the same model to score human texts and model generations and get the sequence-level and token-level negative log-likelihoods. Figure~\ref{fig:neg_log_likelihood} shows the histograms of these negative log-likelihoods.  

In Figure~\ref{fig:neg_log_likelihood}, we take the human text histogram (in blue) as a proxy of \emph{human distribution} and the sampling text histogram (in red) as a proxy of \emph{model distribution}. As you can see, the support of model distribution usually contains the support of human distribution. It supports our previous claim that MLE-trained models tend to over-generalize. 
Meanwhile, at both the sequence and the token levels, the model on average assigns a higher probability to human text than to text sampled from the model.
Therefore, when we promote high-probability sequences or tokens, it is equivalent to pushing the model distribution toward the human distribution. However, we need to avoid overly pushing it to the extremely high-probability region where greedy search outputs locate (in yellow) because they are known to be poor-quality and repetitive. Also, as shown in the figure,  when promoting high-probability \emph{sequences}, even if we overdo it, it will still be within the support of human distribution. In contrast, when promoting high-probability \emph{tokens}, it can go outside the support of the human distribution, which is the drawback of Equation (\ref{eq:final}) compared to Equation (\ref{eq:unfold}).

Lastly, if we train the model only with the self-reinforced objective till convergence, it is inevitable to end up with a model that can only output greedy search generations. Hence, we need to combine it with the forward cross-entropy.

\begin{figure}[t]
    \centering
    \includegraphics[width=0.4\textwidth]{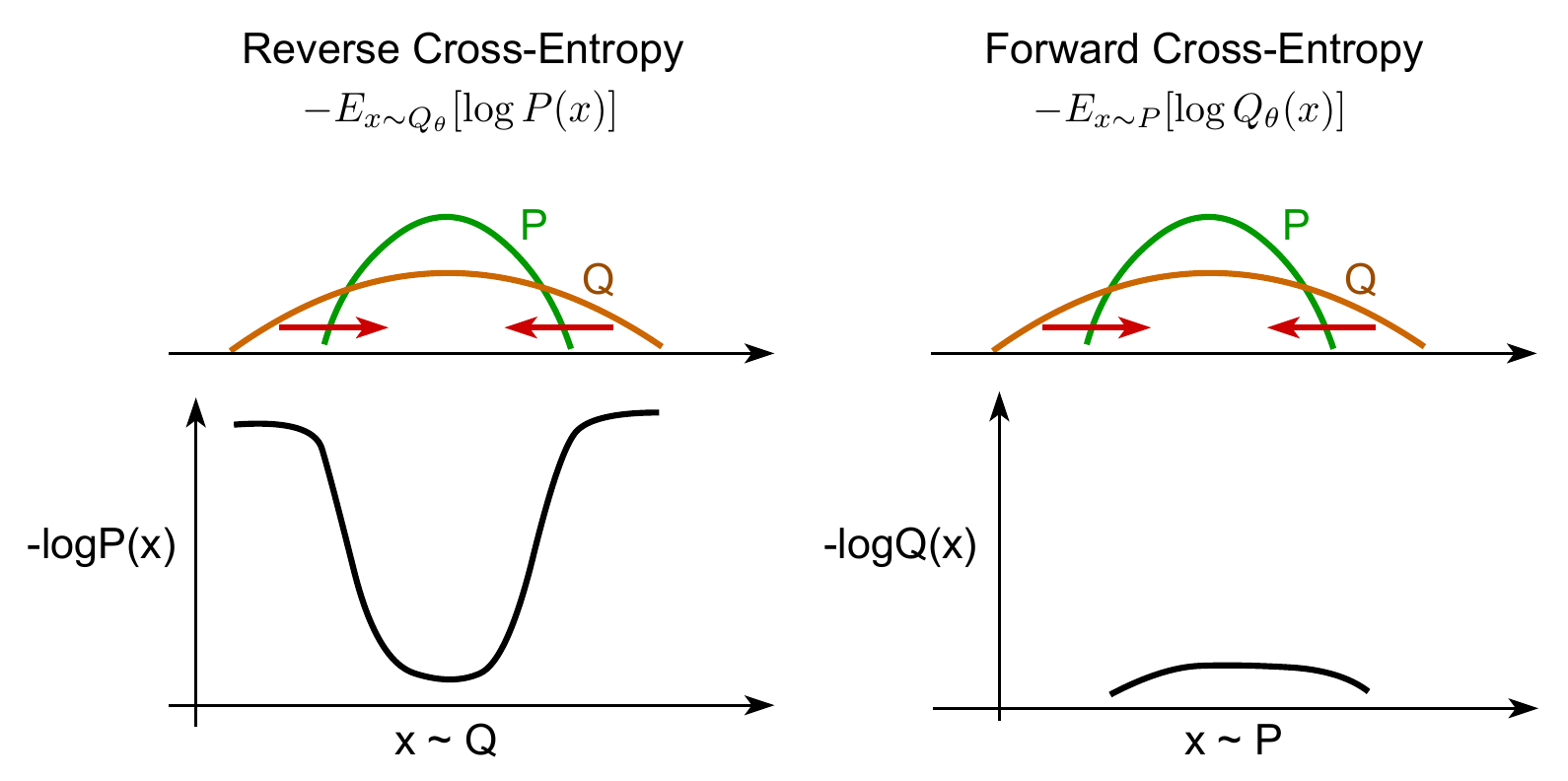}
    \vspace{-5pt}
    \caption{Forward CE only weakly penalizes the model $Q_\theta$ when it puts a small amount of probability mass onto $P(x)=0$ space. And the loss magnitude is much smaller than what we will get from reverse CE.}
    \vspace{-5pt}
    \label{fig:mixces-app}
\end{figure}

\begin{figure*}[t]
    \centering
    \includegraphics[width=0.9\textwidth]{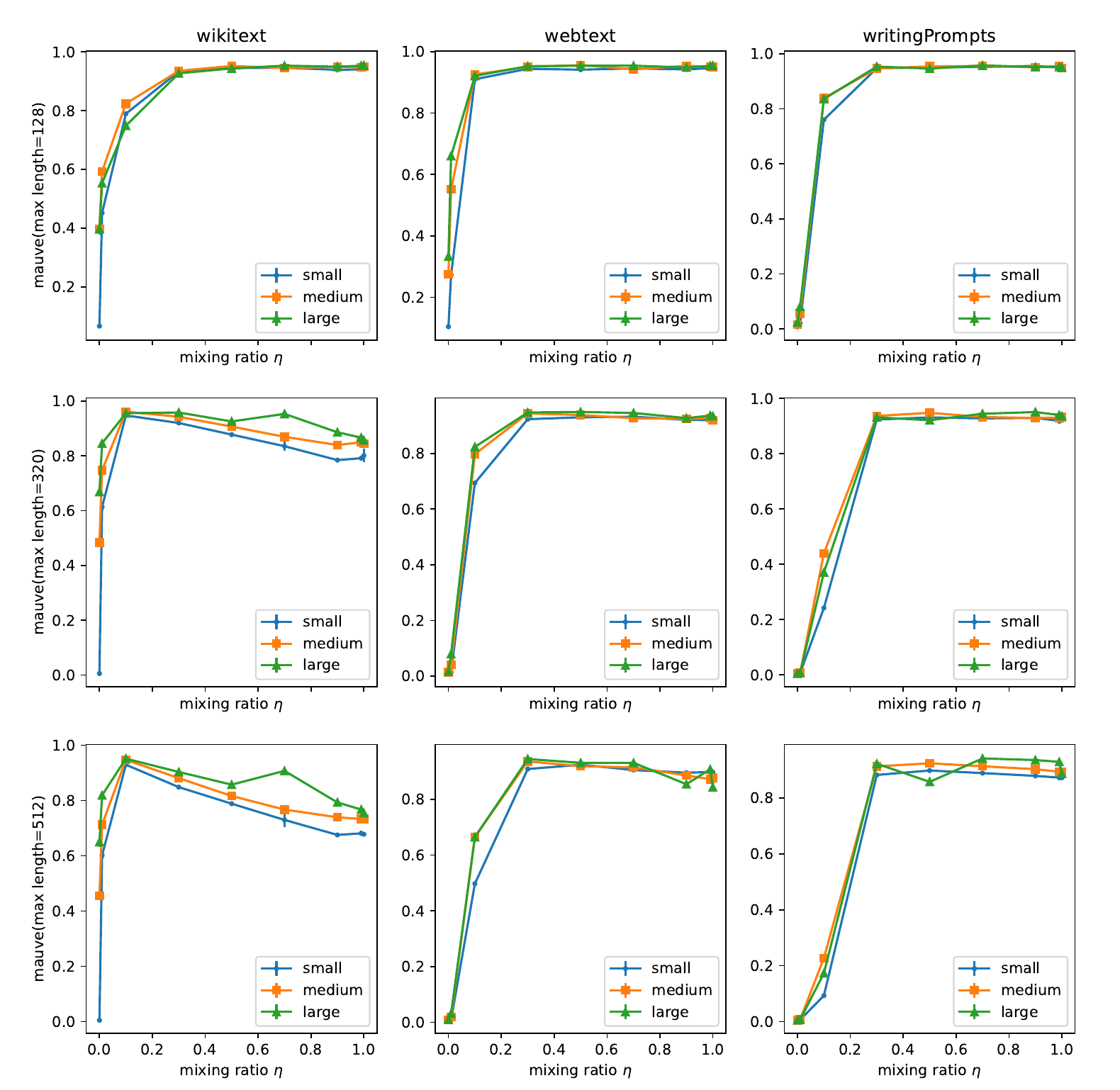}
    \caption{The mauve scores obtained by \mixces{}-finetuned GPT-2 models on development sets with different max generation lengths and different $\eta$.
    Note that when $\eta=1$, \mixces{} is equivalent to MLE. The x-axis is the mixing ratio $\eta$, and the y-axis refers to mauve scores with different max generation lengths. The 3 lines in each subplot show the results of GPT-2 models in different sizes. The 3 subplots in each row are the results of 3 datasets respectively. Unbiased sampling is used as the decoding method. Each dot is the average of 3 runs of sampling and the error bar shows the standard deviation of 3 runs.}
    \label{fig:res_dev}
\end{figure*}

\section{Loss Magnitude}
\label{app:for-rev}
As shown in Figure~\ref{fig:mixces}, we use reverse cross-entropy (CE) to provide a driving force for narrowing the model distribution down when it is broader than the data distribution. And forward CE is to broaden the model distribution up. However, it does not mean forward CE does not have the opposite drive force because forward CE is minimized if and only if $Q_\theta(x) = P(x)$. However, as shown in Figure~\ref{fig:mixces-app}, the loss magnitude is greatly smaller than the loss magnitude we get from reverse CE.

\section{Additional Results}
\subsection{Additional synthethic experiments}

\begin{table}[t]
\centering
\small
\resizebox{0.48\textwidth}{!}{%
\begin{tabular}{ll|cc|cc}
\toprule
& & \multicolumn{2}{c}{Random (10\%)} & \multicolumn{2}{c}{Random (90\%)} \\
\cmidrule(lr){3-4} \cmidrule(lr){5-6} 
Vocab & Objective &  avg. js & avg. 0s  &  avg. js & avg. 0s \\
\midrule
& Gold & 0.0 & 0.0 & 0.0 & 0.0 \\
\midrule
20 & For. KL & 3.65e-4 & 1.80e-4 & 7.56e-4& 9.10e-5 \\ 
& Rev. KL & 3.41e-3 & \bf 5.56e-6 & 1.87e-1& \bf 1.54e-6  \\
& Mix KLs & \bf 3.11e-4& 7.11e-5 & \bf 4.01e-4& 2.67e-5   \\
& JS & 5.68e-3 & 1.17e-5 & 2.14e-1&5.24e-4  \\
& \mixces* & 4.92e-4 & 1.59e-4  & 4.87e-4&2.95e-5 \\
& \mixces & 3.31e-4 & 1.57e-4 & 7.08e-4&8.49e-5 \\
\midrule
50 & For. KL & 6.01e-3 & 1.21e-3 & 2.18e-3&8.90e-5 \\ 
& Rev. KL & 2.03e-2 & \bf 2.01e-5  & 4.11e-1& \bf 4.55e-6 \\
& Mix KLs & \bf 4.65e-3 & 1.29e-4 & 1.54e-3&3.41e-5  \\
& JS & 1.03e-1& 9.03-5 & 4.24e-1&1.25e-5  \\
& \mixces* & 5.20e-3 & 6.84e-4 & \bf 1.48e-3&2.70e-5 \\
& \mixces & 5.96e-3 & 1.20e-3  & 2.03e-3&7.70e-5  \\
\midrule
100 & For. KL & 3.34e-2 & 2.49e-3 & 6.98e-3 & 1.49e-4  \\ 
& Rev. KL & 2.30e-1 & 1.79e-3 & 5.30e-1& \bf 6.25e-6 \\
& Mix KLs & \bf 2.98e-2 & \bf 4.66e-4 & \bf 5.04e-3&6.34e-5 \\
& JS & 2.38e-1  & 1.06e-3  & 5.18e-1&1.32e-3  \\
& \mixces* & 3.10e-2 & 1.73e-3 & 5.12e-3&6.00e-5 \\
& \mixces & 3.29e-2 & 2.44e-3  &7.01e-3&1.50e-5 \\
\midrule
500 & For. KL & 1.56e-1 & 1.57e-3  & 1.93e-1 & 8.45e-4 \\ 
& Rev. KL & 2.94e-1 & 9.91e-4 & 6.49e-1& \bf 2.33e-6 \\
& Mix KLs & \bf 1.55e-1 & 1.45e-3  & 1.70e-1&6.83e-4 \\
& JS & 2.95e-1& \bf 9.78e-4  & 5.75e-1&1.35e-3 \\
& \mixces* & \bf 1.55e-1& 1.45e-3  & \bf 1.69e-1&6.71e-4  \\
& \mixces &\bf 1.55e-1 & 1.56e-3  & 1.88e-1 & 6.28e-4 \\
\midrule
1000 & For. KL & 1.83e-1& 8.95e-4 & 3.65e-1& 7.31e-4\\ 
& Rev. KL & 2.86e-1 & 6.12e-4  & 6.68e-1&\bf 3.88e-6  \\
& Mix KLs & \bf 1.80e-1 & 8.64e-4 & 3.50e-1&6.86e-4 \\
& JS & 2.88e-1 & \bf 6.11e-4 & 5.80e-1&7.73e-4 \\
& \mixces* & 1.83e-1 & 8.64e-4 & 3.50e-1&6.84e-4 \\
& \mixces & 1.83e-1 & 8.92e-4  & \bf 3.48e-1&6.71e-4  \\
\bottomrule
 \end{tabular}
 }
 \vspace{-5pt}
\caption{ The results of the other two synthetic experiments. Random (10\%) and Random (90\%) both use random initialization for $\mathbf{M}$, and 10\% and 90\% probabilities in $\mathbf{M}$ are 0 respectively. Gold refers to the results when $\mathbf{M}'$=$\mathbf{M}$. Each value is a 5-seed average.} 
\label{table:syn-app}
\vspace{-5pt}
\end{table}

Table~\ref{table:syn-app} shows the results of additional synthetic experiments besides Table~\ref{table:syn} in the main paper. Here, the goal transition matrix $\mathbf{M}$ is randomly initialized with 10\% and 90\% zero probabilities.

As the magnitudes of both avg. js and avg. 0s are fairly small, we examine the 95\% confidence intervals under one synthetic experimental setting -- initializing the transition matrix $\mathbf{M}$ by the bigram occurrence in the WebText data and setting vocabulary size as 1000. Table~\ref{table:ci} contains the results. We can see that 95\% confidence intervals are small enough to maintain the trend of the results. 

\begin{table}[t]
\centering
\small
\resizebox{0.48\textwidth}{!}{%
\begin{tabular}{ll|cc}
\toprule
& & \multicolumn{2}{c}{WebText} \\
\cmidrule(lr){3-4} 
Vocab & Objective & avg. js & avg. 0s \\
\midrule
1000 & For. KL & 8.10e-2 $\pm$ 2.45e-4 & 1.50e-4 $\pm$ 5.58e-7\\ 
& \mixces* & \textbf{7.44e-2} $\pm$ 2.46e-4 & \textbf{1.14e-4} $\pm$ 6.15e-7\\
& \mixces & 7.94e-2 $\pm$ 2.15e-4 & 1.42e-4 $\pm$ 5.05e-7 \\
\bottomrule
 \end{tabular}
 }
 \vspace{-5pt}
\caption{Synthetic experimental results with 95\% confidence intervals. WebText means initializing $\mathbf{M}$ by the bigram occurrence in the WebText data.} 
\label{table:ci}
\vspace{-5pt}
\end{table}

\subsection{Varying training data sizes}
Table~\ref{table:data-size} shows the results of using different training data sizes in the real-data setting. 

\subsection{Varying $\eta$ and max generation length}
Figure~\ref{fig:res_dev} illustrates the curves of mauve scores on the development sets.

\begin{table*}[t]
\centering
\small
\resizebox{0.98\textwidth}{!}{%
\begin{tabular}{ll|c|ccc|ccc}
\toprule
& & WikiText & \multicolumn{3}{c}{WebText} & \multicolumn{3}{c}{WritingPrompts} \\
\cmidrule(lr){3-3} \cmidrule(lr){4-6} \cmidrule(lr){7-9} 
Model Size & Objective & c-mauve$_{100}$ & c-mauve$_{100}$ &  c-mauve$_{200}$ & c-mauve$_{300}$ & c-mauve$_{100}$ &  c-mauve$_{200}$ & c-mauve$_{300}$\\
\midrule
& Human & 0.97 & 0.96  & 0.96 & 0.96 & 0.96 & 0.96 & 0.96  \\
\midrule
\multirow{2}{*}{Small} & MLE & \bf 0.92 & 0.93 & 0.92 & 0.90 & 0.94 & \bf 0.94 & 0.92 \\
& \mixces & \bf 0.92 & \bf 0.94 & \bf 0.94 & \bf 0.93 & \bf 0.95 &\bf 0.94 & \bf 0.94 \\
\midrule
\multirow{2}{*}{medium} & MLE & \bf 0.94 & 0.93 & 0.91 & 0.90 & 0.94 & \bf 0.94 & 0.93\\
& \mixces &  0.93 & \bf 0.95 & \bf 0.94 &\bf 0.94 & \bf 0.95 & \bf 0.94 & \bf 0.94 \\
\midrule
\multirow{2}{*}{Large} & MLE &  \bf 0.93 & 0.93 & 0.93 & 0.91 & 0.94 & 0.94 & 0.93 \\
& \mixces & \bf 0.93 & \bf 0.94 & \bf 0.94 &\bf  0.93 & \bf 0.95 & \bf 0.95 & \bf 0.95 \\
\bottomrule
 \end{tabular}
 }
 \vspace{-5pt}
\caption{Controlled mauve results. Unbiased sampling is used as the decoding method, i.e., using the same model generations as Table~\ref{table:model-size}. Human scores are not 1 because sampling 10K fragments twice result in two different sets. Each number is a 3-run average. } 
\label{table:cotrolled-mavue}
\end{table*}

\begin{table*}[t]
\centering
\small
\resizebox{0.82\textwidth}{!}{%
\begin{tabular}{ll|c|ccc|ccc}
\toprule
& & WikiText & \multicolumn{3}{c}{WebText} & \multicolumn{3}{c}{WritingPrompts} \\
\cmidrule(lr){3-3} \cmidrule(lr){4-6} \cmidrule(lr){7-9} 
Model Size & & c-coh$_{100}$ & c-coh$_{100}$ &  c-coh$_{200}$ & c-coh$_{300}$ & c-coh$_{100}$ &  c-coh$_{200}$ & c-coh$_{300}$\\
\midrule
& Human &  0.570 & 0.521 & 0.583 &  0.600 & 0.412 & 0.470 & 0.481\\
\midrule
\multirow{2}{*}{Small} & MLE & 0.504 & 0.444 & 0.515 & 0.535 & 0.350 & 0.412 & 0.429\\
& \mixces &\bf 0.508 & \bf 0.458 & \bf 0.524 & \bf 0.545 & \bf 0.363 & \bf 0.422 & \bf 0.437\\
\midrule
\multirow{2}{*}{Medium} & MLE & 0.518 & 0.446 & 0.515 & 0.535 & 0.355 & 0.415 & 0.432\\
& \mixces & \bf 0.527 & \bf 0.484 & \bf 0.546 & \bf 0.565 & \bf 0.362 & \bf 0.425 & \bf 0.437\\
\midrule
\multirow{2}{*}{Large} & MLE & 0.521 & 0.449 & 0.515 & 0.536 & \bf 0.372& 0.431 & 0.447\\
& \mixces & \bf 0.522 & \bf 0.469 & \bf 0.531 & \bf 0.569 & 0.369 & \bf 0.434 & \bf 0.450 \\
\bottomrule
 \end{tabular}
 }
 \vspace{-5pt}
\caption{Controlled coherence results. Unbiased sampling is used as the decoding method, i.e., using the same model generations as Table~\ref{table:model-size}. Each number is a 3-run average. } 
\label{table:cotrolled-coh}
\vspace{-5pt}
\end{table*}

\subsection{Controlled Mauve and Coherence}
\label{app:controlled-coh}
We find that the actual length of the text is a confounding factor of mauve computation. For example, when we compute mauve between a set of texts and the same set with an extra new line token after each text (or the same set with the last k tokens being truncated), the score will be lower than 0.01. Though you may think truncating all texts to the same length can resolve this problem, we find that the \emph{incompleteness} caused by truncation can also be a confounding factor. For instance, keeping human texts intact, we truncate texts generated by two systems by their shorter lengths (i.e., for each example, we truncate text1 and text2 by min\_length(text1, text2)). Then, the system whose texts get truncated less will get a greatly larger mauve score than the other system. Therefore, to eliminate the influence of these two confounding factors, we propose a \emph{controlled mauve} computation approach. Concretely, for the set of human texts $\mathbf{T}_h$ and the set of model-generated texts $\mathbf{T}_m$, we randomly sample 10K $L$-length text fragments from each of these two sets. $L$ is the number of tokens in each text fragment. After that, we compute the mauve between these two sets of 10K text fragments. We denote this controlled mauve as c-mauve$_L$.
$$\mathbf{F}_{h, L} = \{f_{h, L}^i\}_{i=1}^{10K}, f_{h, L}^i \sim \mathbf{T}_h$$
$$\mathbf{F}_{m, L} = \{f_{m, L}^i\}_{i=1}^{10K}, f_{m, L}^i \sim \mathbf{T}_m$$
$$\text{c-mauve}_L = \text{mauve}(\mathbf{F}_{h, L}, \mathbf{F}_{m, L})$$
To sample each fragment, we first randomly sample a text $t^i$ from the set, and then randomly select a start token $s$ (as long as there are more than $L$ tokens from $s$ to the end of $t^i$), then the fragment is $t^i[s:s+L]$. 
Finally, Table~\ref{table:cotrolled-mavue} shows the results. We set $L=$ 100, 200, and 300, except that we could not get 10K 200-token fragments from WikiText because its texts are shorter. 

The Coherence score~\cite{su2022contrastive} computes the cosine similarity between the prompt and the continuation. We suspect that the length of the continuation may affect the score. Therefore, following the same idea of controlled mauve, we also sample 10K fragments of the same length from the set of texts for evaluation and compute coherence on the fragments. And for each fragment, we take the first 50 tokens as the prompt and the rest as the continuation. Table~\ref{table:cotrolled-coh} shows the results. As you can observe, under this controlled setting, \mixces{}-finetuned models generally achieve better coherence over MLE-finetuned models. 

\subsection{Text length of model generations}

\begin{table}[t]
\centering
\small
\resizebox{0.48\textwidth}{!}{%
\begin{tabular}{ll|c|c|c}
\toprule
& & WikiText & WebText & WritingPrompts \\
\cmidrule(lr){3-3} \cmidrule(lr){4-4} \cmidrule(lr){5-5} 
Model Size & Objective &  avg. len &  avg. len &  avg. len \\
\midrule
& Human & 124.5 & 304.5 & 332.5  \\
\midrule
\multirow{2}{*}{Large} & MLE & 114.8 & 284.2 & 325.8  \\
& \mixces &  89.0 & 298.9 & 326.4 \\
\bottomrule
 \end{tabular}
 }
 \vspace{-5pt}
\caption{Unbiased sampling text lengths of models finetuned by MLE or \mixces{} on three datasets. Length is computed by simply splitting text by whitespaces.} 
\label{table:text_len}
\vspace{-5pt}
\end{table}

Though by default we set the max generation length as 512, the actual text length can vary as the EOS token can be sampled at any time step. Therefore, we list the average text length of the human text and GPT2-large generations in Table~\ref{table:text_len}. We observe that model generations are always shorter than human text. Compared to MLE, our \mixces{}-finetuend model produces shorter text on WikiText while producing longer text on the other two datasets. We suspect that the shorter length of \mixces{} on WikiText is due to the small mixing ratio (0.1) chosen based on mauve (see Table~\ref{table:gpt2-eta}). However, we do not think shorter text length leaves to better mauve, as shown by the other two datasets and discussed in~\ref{app:controlled-coh}.

\begin{table*}[t]
\centering
\small
\begin{tabular}{ll|c|c|c|c}
\toprule
\multicolumn{6}{l}{Model section is based on avg. js} \\
\midrule
& & Random (50\%) & WebText & Random (10\%) & Random (90\%)\\
\cmidrule(lr){3-3} \cmidrule(lr){4-4}  \cmidrule(lr){5-5}  \cmidrule(lr){6-6}
Vocab & Objective &  best $\eta$  &  best $\eta$  &  best $\eta$  &  best $\eta$\\
\midrule
20 & Mix KLs &  0.99 & 0.9 & 0.99 & 0.99 \\
& JS &  0.9 & 0.9 & 0.9& 0.9\\
& \mixces* & 0.99 & 0.99 & 0.99& 0.99\\
& \mixces & 0.9 & 0.99 & 0.99& 0.99\\
\midrule
50 & Mix KLs & 0.99 & 0.99 & 0.9& 0.99\\
& JS & 0.01 & 0.99 & 0.9& 0.9\\
& \mixces* & 0.99 & 0.99 & 0.99& 0.99\\
& \mixces &  0.99 & 0.99 & 0.99& 0.9\\
\midrule
100 & Mix KLs & 0.9 & 0.99 & 0.9& 0.99\\
& JS & 0.01 & 0.99 & 0.99& 0.01\\
& \mixces* & 0.99 & 0.99 & 0.99& 0.99\\
& \mixces & 0.5 & 0.9 & 0.5& 0.99\\
\midrule
500 & Mix KLs & 0.9 & 0.99 & 0.99& 0.99\\
& JS & 0.99 & 0.99 & 0.99& 0.99\\
& \mixces* & 0.99 & 0.99 & 0.99& 0.99\\
& \mixces & 0.1 & 0.5 & 0.1& 0.1\\
\midrule
1000 & Mix KLs & 0.99 & 0.99 & 0.99 & 0.99 \\
& JS & 0.99 & 0.99& 0.99 & 0.99\\
& \mixces* & 0.99 & 0.99& 0.99 & 0.99\\
& \mixces & 0.1 & 0.5 & 0.1& 0.1\\
\bottomrule
 \end{tabular}
 \vspace{-5pt}
\caption{The selected best $\eta$ of synthetic experiments reported in Table~\ref{table:syn} and Table~\ref{table:syn-app}. The model section is based on avg. js.} 
\label{table:syn-eta}
\end{table*}

\begin{table*}[t]
\centering
\small
\begin{tabular}{ll|c|c|c}
\toprule
\multicolumn{5}{l}{Model section is based on mauve (max length=512) on dev set} \\
\midrule
& & WikiText & WebText & WritingPrompts \\
\cmidrule(lr){3-3} \cmidrule(lr){4-4} \cmidrule(lr){5-5} 
Model Size & Objective & best $\eta$ & best $\eta$ &  best $\eta$ \\
\midrule
Small & \mixces & 0.1 & 0.5 & 0.5\\
\midrule
Medium & \mixces & 0.1 & 0.3 & 0.5 \\
\midrule
Large & \mixces & 0.1 & 0.3 & 0.7\\
\bottomrule
 \end{tabular}
 \vspace{-5pt}
\caption{The selected best $\eta$ of GPT-2 experiments reported in Table~\ref{table:model-size}. The model section is based on mauve (max length=512) on the dev set. } 
\label{table:gpt2-eta}
\vspace{-5pt}
\end{table*}

\section{Best $\eta$s}
Table~\ref{table:syn-eta} has the best $\eta$s for synthetic experiments.
Table~\ref{table:gpt2-eta} contains the best $\eta$s selected for GPT-2 experiments.

\begin{table*}[t]
\centering
\small
\begin{tabular}{l|c|c|c}
\toprule
Dataset& all agree & 2 agree & no agreement \\
\midrule
WikiText & 24\% & 59\% & 17\%\\
\midrule
WebText & 24\% & 52\% & 24\% \\
\midrule
WritingPrompts & 20\% & 70\% & 10\% \\
\bottomrule
 \end{tabular}
 \vspace{-5pt}
\caption{Inter-annotator agreement. The numbers are the portions of examples that have a 3-annotator agreement (all agree), a 2-annotator agreement (2 agree), or no agreement. E.g., 24\% of examples used in human evaluation for WikiText have an agreement among 3 annotators.} 
\label{table:agreement}
\vspace{-5pt}
\end{table*}

\section{Human Evaluation Details}
\label{app:human_eval}
\begin{figure*}
    \centering
    \includegraphics[width=0.9\textwidth]{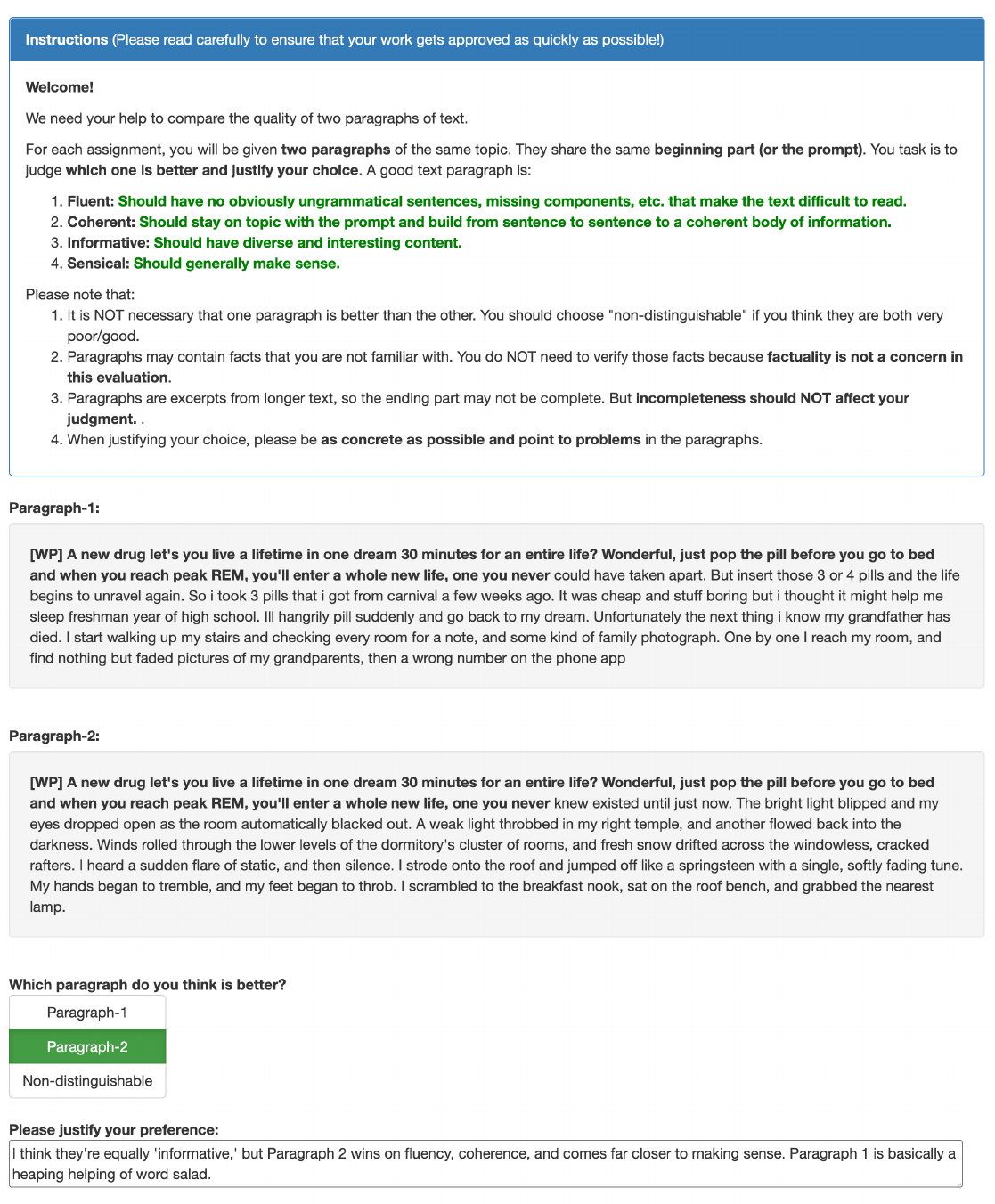}
    \caption{Human evaluation interface and a random example from our collected human annotations.}
    \label{fig:mturk}
\end{figure*}

We conduct A/B testing (or pairwise comparison) to compare generations from two models. As shown in Figure~\ref{fig:mturk}, in each job, we give the evaluator two text paragraphs (in random order) that share the same beginning part (the prompt) but have different continuations. Then, they need to choose which one they think is better (or non-distinguishable). To avoid random selections, they are also asked to provide a justification for their choice. We find this justification not only gives us additional explanations of their choices but also helps us easily identify bad workers, because bad workers tend to use one single justification or several repeated justifications. 

We instruct them by defining a good text paragraph as being:
\begin{itemize}
    \item \textbf{Fluent}: Should have no obviously
ungrammatical sentences, missing components, etc. that make the text difficult to read.
    \item \textbf{Coherent}: Should stay on topic with the prompt and build
from sentence to sentence to a coherent body of information.
    \item \textbf{Informative}: Should have diverse and interesting content.
    \item \textbf{Sensical:} Should generally make sense.
\end{itemize}

Since short text has little information and long text is difficult to read, we only use paragraphs with 5 to 8 sentences for evaluation. If a paragraph has more than 8 sentences, we truncate it to 8 sentences. And we remove paragraphs with less than 400 or more than 2000 characters. Besides, to eliminate the influence of length difference, we do not select examples whose length difference between two paragraphs is more than 1 sentence or more than 200 characters. 

We conduct this evaluation on Amazon Mechanical Turk. We only allow workers, who are located in the US, have a Masters Qualification,\footnote{\url{https://www.mturk.com/worker/help}} have an approval rate larger than 97\%, and have more than 10000 HITs approved, to do our tasks. In addition, we first ran a testing batch, then manually checked the results, and selected 44 qualified workers to continue doing the rest of our tasks. 

For each of the 3 datasets, we sampled 105 examples and collected 3 responses per example. In total, we received 945 human evaluations. We pay workers \$1 per response, and it takes around 5 minutes to finish one response, i.e., the hourly rate is around \$12. 

Table~\ref{table:agreement} shows that inter-annotator agreements. Figure~\ref{fig:wikitext-example1}-\ref{fig:writingprompts-example2} are 6 randomly sampled examples from human evaluation results, 2 examples per dataset.

\section{Reproducibility}
\label{app:repro}
In our GPT-2 experiments, we use English text data from 3 domains: (1) WikiText~\cite{merity2016pointer}: text from Wikipedia, and we use wikitext-103-raw-v1 from Hugging Face.\footnote{\url{https://huggingface.co/datasets/wikitext}} Its license is Creative Commons Attribution-ShareAlike License (CC BY-SA 4.0). (2) WebText~\cite{radford2019language}: text from the Web. It was used for pretraining GPT-2. The full WebText is not available but they released a subset on Github\footnote{\url{https://github.com/openai/gpt-2-output-dataset}}. The GitHub repository contains an MIT license, and they did not specify the license of the data. But they indicated in the readme: ``We look forward to the research produced using this data!'' (3) WritingPrompts~\cite{fan-etal-2018-hierarchical}\footnote{\url{https://github.com/facebookresearch/fairseq/tree/main/examples/stories}}: text from the writing prompts forum of Reddit. Its GitHub repository also contains an MIT license without specification of the data license. However, WritingPrompts has been used by many other research works, e.g., ~\citet{pillutla2021mauve}. We use their official dev and test sets as much as possible. If they have fewer than 5K examples, we sample from their official training set to make up the rest. 

All of our experiments were conducted on NVIDIA Tesla V100 32G GPUs. We use a single GPU to run each experiment and change the batch size to fit models of different sizes. When fine-tuning GPT-2 small using a single GPU with MLE or \mixces{}, it took less than 1 hour to finish 5 epochs on 50K WikiText training data and took less than 2 hours to finish 5 epochs on 50K WebText or WringPrompts training data. 

We implemented our GPT-2 based models based on the GPT-2 modeling code from Hugging Face Transformers\footnote{\url{https://github.com/huggingface/transformers/blob/main/src/transformers/models/gpt2/modeling_gpt2.py}}. For training and evaluation, we modified the example script of causal language model training\footnote{\url{https://github.com/huggingface/transformers/blob/main/examples/pytorch/language-modeling/run_clm_no_trainer.py}}. We used the default optimizer, learning rate, scheduler, etc. in that script. But we set the maximum training epochs as 5 and changed the batch size and gradient accumulation steps based on the model size to fit it in one 32G-memory GPU.

\begin{figure*}[ht]
\begin{center}
\small
\begin{tabularx}{\textwidth}{X}
\toprule 
\textbf{Paragraph1 (MLE)}: \\
\textbf{Within minutes of issuing the signal and turning his flagship HMS Queen Charlotte, Howe's plan began to falter. Many of the British captains had either misunderstood or ignored the signal and were hanging back in the original line. Other ships were still struggling with damage to their} ships and other equipment caused by the storm, so that Australia's war effort was threatened. In response to the storm-ravaged Australian distress call on 12 March, Howe ordered his ship HMS Sun Babies as flagship of rowing party V HMNZS Platypus. He assigned elevensacks from the ship to the crew for a rescue, and placed much of the more aged of his crew and stockmen on sick leave on the advice of Admiral Rear-Admiral John Forbes.\\ \\
\textbf{Paragraph2 (\mixces{})}: \\
\textbf{Within minutes of issuing the signal and turning his flagship HMS Queen Charlotte, Howe\'s plan began to falter. Many of the British captains had either misunderstood or ignored the signal and were hanging back in the original line. Other ships were still struggling with damage to their} yards and were standing still. Another officer, Lieutenant Colonel Strother, privately stated that the British ship captains had "obsessed [ed] the whole thing over with the consciousness of the impossibility of victory". Commander Howe was eager to proclaim victory, but it was not until the Chief of the General Staff, Sir Charles Ogilvy, finally appeared on the scene that he decided to make a decision. \\
\midrule
\textbf{Choice}: paragraph2 \\
\textbf{Justification}: Paragraph 1 has details that do not flow with each other as if Howe was involved with two ships at the same time. The beginning of the paragraph mentions Howe's flagship HMS Queen Charlotte and later on his flagship HMS Sun Babies. It is not clear about the role of Australians whether there were Australians in the crew or they were helping Australian ships. In paragraph 2, there is no mention of another nation other than just British. It has details of different ranks of commands. Moreover the paragraph focus on failure and uncertainty that flow to the rest of the paragraph. \\ \\

\textbf{Choice}: paragraph2 \\
\textbf{Justification}: Paragraph 2 seems to make more sense and is more coherent than Paragraph 1. \#2 has a narrative flow that can be read from start to finish, giving the reader a sense of what is happening. The quote is the weak point, but as a quote, can be taken as it is given here. On the other hand, Paragraph 1 can also be read from start to finish, but makes less sense, and seems less coherent. \\ \\

\textbf{Choice}: paragraph2 \\
\textbf{Justification}: Paragraph 2 is more coherent, and it wins hands down in terms of making sense. Paragraph 1 has Howe sending men on leave in the middle of a battle and also has him assigning "elevensacks" to the crew, which isn't even a real word. \\
\bottomrule
\end{tabularx}
\end{center}
\vspace{-10pt}
\caption{WikiText example 1.}
\vspace{-12pt}
\label{fig:wikitext-example1}
\end{figure*}

\begin{figure*}[ht]
\begin{center}
\small
\begin{tabularx}{\textwidth}{X}
\toprule 
\textbf{Paragraph1 (MLE)}: \\
\textbf{In imitation of the Greeks, Nero built a number of gymnasiums and theatres. Enormous gladiatorial shows were also held. Nero also established the quinquennial Neronia. The festival included games, poetry, and theater.} For the recesses of the Calendrical Mysterium, the city quinquenored, with the city many times superseded.[ndx] Nero also Festina-Illuminatae in honour of the doomed Emperor Caeser and his son Severus, the founders of the Flavian dynasty, who were slain in that period. In Britain, Nero erected the Attic theatre.\\ \\
\textbf{Paragraph2 (\mixces{})}: \\
\textbf{In imitation of the Greeks, Nero built a number of gymnasiums and theatres. Enormous gladiatorial shows were also held. Nero also established the quinquennial Neronia. The festival included games, poetry, and theater.} In order to combat the influence of the Venetian party, the local aristocracy discouraged the shooting of arrows at the official spectators. Nero's penchant for shooting arrows became a local insult.\\
\midrule
\textbf{Choice}: paragraph2 \\
\textbf{Justification}: The first one is hard to understand. What does " the city quinquenored, with the city many times superseded.[ndx]" mean.? It does not make sense. The second one is easy to understand. \\ \\

\textbf{Choice}: paragraph2 \\
\textbf{Justification}: I had to read Paragraph 1 a few times before it made sense, but Paragraph 2 flowed nicely and logically without effort. \\ \\

\textbf{Choice}: paragraph2 \\
\textbf{Justification}: 2 takes a bit of a stretch but it can at least follow some plausible logical order. 1 just abruptly starts talking about geopolitics which doesn't make sense in that context. \\
\bottomrule
\end{tabularx}
\end{center}
\vspace{-10pt}
\caption{WikiText example 2.}
\vspace{-12pt}
\label{fig:wikitext-example2}
\end{figure*}

\begin{figure*}[ht]
\begin{center}
\small
\begin{tabularx}{\textwidth}{X}
\toprule 
\textbf{Paragraph1 (MLE)}: \\
\textbf{Two years ago, Sandrine Jourdan walked up a friend's Sunshine Coast driveway and simply disappeared. The Caboolture woman has not accessed her bank accounts, used her phone or wished any of her three children a happy birthday since then.} She insists she hasn't moved house, childcare is still open and she can speak to her family now. "I owe them no money." Jourdan said. Jourdan was last heard from around February this year and her family does not know where she is. One of her family members told me Jourdan was hers but to leave Jourdan alone would be like asking the sick to go to the doctor. However, the woman, who is in her 50s, said Jourdan was extremely important to her family and they didn't want to send her any money. They saw her daily and had almost no tensions of any kind to handle\\ \\
\textbf{Paragraph2 (\mixces{})}: \\
\textbf{Two years ago, Sandrine Jourdan walked up a friend's Sunshine Coast driveway and simply disappeared. The Caboolture woman has not accessed her bank accounts, used her phone or wished any of her three children a happy birthday since then.} But she has no one to blame but herself. The welfare office at Taree hostel was "terminating her accommodation on a whim" last Spring, she says. Discarded money The hostel is not Charter and cannot employ staff. Desperate for a fresh start, Sandrine tried to access her bank accounts and managed to access her daughter's allowance, the bequests of her parents, a benefit card and money in savings. She has not seen her family since. To compound her plight, she has lost her job on the condominium project near the town of Machalung where she was staying \\
\midrule
\textbf{Choice}: non-distinguishable \\
\textbf{Justification}: Both are bad and neither follow the prompt. Paragraph 1 isn't coherent. She is on good terms with her family, then not, then back again. Paragraph 2 also has contradictions, and the sentence "Discarded money The hostel is not Charter and cannot employ staff. " seems randomly inserted.\\ \\

\textbf{Choice}: non-distinguishable \\
\textbf{Justification}: Neither follows the prompt in the least.  It is about a woman who disappears, and both go on to have quotes and information about the missing woman as if she were still interacting with others. There are incomplete sentences in 2 which are not present in 1, but I still can't say 1 is better because it's so off prompt. \\ \\

\textbf{Choice}: paragraph1 \\
\textbf{Justification}: Paragraph 2 has some punctuation and grammar errors, for example "Discarded money The hostel is not Charter and cannot employ staff." \\
\bottomrule
\end{tabularx}
\end{center}
\vspace{-12pt}
\caption{WebText example 1.}
\vspace{-10pt}
\label{fig:webtext-example1}
\end{figure*}

\begin{figure*}[ht]
\begin{center}
\small
\begin{tabularx}{\textwidth}{X}
\toprule 
\textbf{Paragraph1 (MLE)}: \\
\textbf{Another Phantom is leaving the building. The musical, which is the longest running show in Broadway history, announced Monday that the Phantom, James Barbour, will leave the production on Dec. 23, 2017, after a close to three-year run.} A premature announcement about the closure came shortly after the show was first announced at the 2016 Tony Awards. A representative for JBL had told The Hollywood Reporter's Hayden Dingman last year that the Broadway appearance would likely wrap but that there were still plans to continue performing on the stage. The event marked the end of the run of 'The Phantom' which reached a two-decade high in the Broadway season. The show had run a whopping 129 shows on Broadway since its inception in 1995. After 11 seasons, Ice Cube's 'Once', which had been announced 14 months prior, was a box office success \\ \\
\textbf{Paragraph2 (\mixces{})}: \\
\textbf{Another Phantom is leaving the building. The musical, which is the longest running show in Broadway history, announced Monday that the Phantom, James Barbour, will leave the production on Dec. 23, 2017, after a close to three-year run.} "Moving on to a new chapter with a new partner is one of the hardest things we've ever done as a show," said producer Scott Ploog. "Joining forces with our fearless leader James was a huge statement to us." After singing and dancing its way through the Fort Worth Convention Center in 2011, the show was working on a script for its sixth season. The Phantom surprised the show's production staff with a call from James, who was the project's conductor for the years it was running. "After hearing the love he had for the show, I knew whether I wanted to take a job as indispensable leader of the cast or what, I had to decide," said Expedition leader Cary Joji. "He was just engaging, embracing and completely out of his ways, but he was relentless \\
\midrule
\textbf{Choice}: paragraph2 \\
\textbf{Justification}: Paragraph 2 feels more personal with more quotes from the people that worked on the show.\\ \\

\textbf{Choice}: paragraph2 \\
\textbf{Justification}: Both do well on fluency. For coherence, I don't think either one forms a totally coherent body of information, but Paragraph 2 sticks to the topic a touch better. I think they're roughly equal in terms of diverse and interesting content. As usual, making sense is the metric each is worst at. I believe they are indistinguishably terrible in that regard. So Paragraph 2 wins, just barely, via being slightly more coherent. \\ \\

\textbf{Choice}: paragraph2 \\
\textbf{Justification}: The first paragraph seems to be a mix of unrelated items, at first discussing a performer leaving the cast of Phantom and then talking about the end of the production and then concluding with "After 11 seasons, Ice Cube's 'Once', which had been announced 14 months prior, was a box office success", which seems unrelated to Phantom. The second paragraph seemed little off also, but had somewhat more coherence. The first paragraph also lacked a couple comma, by rule, but I didn't factor that in very much. \\
\bottomrule
\end{tabularx}
\end{center}
\vspace{-12pt}
\caption{WebText example 2.}
\vspace{-10pt}
\label{fig:webtext-example2}
\end{figure*}

\begin{figure*}[ht]
\begin{center}
\small
\begin{tabularx}{\textwidth}{X}
\toprule 
\textbf{Paragraph1 (MLE)}: \\
\textbf{[WP] There is no such thing as ``dating '' anymore. Just like in Cinderella, As soon as you touch the hand of`` the one'' you have an insatiable urge to kiss. Sam sighed. Despite her best efforts to} hide it, the bruises and cuts beneath her blouse were pretty apparent from her sitting down. Her heart pierced so many ribs that buckles broke, and chips torn the flesh of her ribs, arm and leg. And they weren't just sitting there - it was pouring out down the side of the castle. When they arrived back at the villa after a short- but exhausting trip from the train station, they were determined to hug and kiss, just as well as they could in dark alleys. Unfortunately, the darkness blinded them from view, so the light was flowing with the pouring rain, sending sparks from their inexpensive outfits\\ \\
\textbf{Paragraph2 (\mixces{})}: \\
\textbf{[WP] There is no such thing as ``dating '' anymore. Just like in Cinderella, As soon as you touch the hand of`` the one'' you have an insatiable urge to kiss. Sam sighed. Despite her best efforts to} reverse the trend, her hand seemed to instinctively take the hand of the lone participant. He smiled, as Sam closed her eyes. "Today's the day you know, Yattar," he insisted confidentially. "Today's the day you know the thrill of kissing!" Yattar nodded, with conviction. "Anyone the purveyors of-" "Remind me to spice everything up," she dismissed, this time making eye contact with the man her mouth had caught them \\
\midrule
\textbf{Choice}: paragraph2 \\
\textbf{Justification}: Assuming that "Yattar" is a pet name/different name for "Sam", 2 is much more cohesive and fluent than 1.  It is easy to follow the story and the story makes more sense than 1.  The "pouring out down the side of the castle" in 1 makes no sense, and other content is pretty incoherent as well.\\ \\

\textbf{Choice}: paragraph1 \\
\textbf{Justification}: Paragraph 1 is a more concrete story of love, while 2 feels more disjointed.\\ \\

\textbf{Choice}: paragraph2 \\
\textbf{Justification}: Paragraph 2 is more cohesive, sticking to the concept introduced in the first sentence and describing the initial meeting of a couple. Para 1 rather jarringly adds a description of injuries and a castle setting. \\
\bottomrule
\end{tabularx}
\end{center}
\vspace{-10pt}
\caption{WritingPrompts example 1.}
\vspace{-12pt}
\label{fig:writingprompts-example1}
\end{figure*}

\begin{figure*}[ht]
\begin{center}
\small
\begin{tabularx}{\textwidth}{X}
\toprule 
\textbf{Paragraph1 (MLE)}: \\
\textbf{[CC] I wrote these stories recently here on WP, and would like to share them with my friends; however, I feel like they could be tuned up a bit. Any thoughts on how to make improvements is greatly appreciated. First off,} there are a lot of inconsistencies in their sentences, it's good to have a prompt where you can color all the immersion steps individually so you can judge meaningful sequences when looking at them in context. 2nd, while I did my best to adhere to game mechanics in the story, they were probably made largely to draw parallels between the reader and protagonist. Seth Vargo Hughes is an unfortunately short person himself (7 weeks a year usa is somewhere between 8 to 10 years long. I amn't entirely sure what his killer skills are now, or about what he can do), so he may get caught with some pretty nasty tools and weapons, and at the end of the story, he's not gonna give the guy who rescued him the satisfaction of just putting his fist through his friend's chest to make his statement true. But that happens. 2nd, 2 other things that PWP failed to properly convert into told story\\ \\
\textbf{Paragraph2 (\mixces{})}: \\
\textbf{[CC] I wrote these stories recently here on WP, and would like to share them with my friends; however, I feel like they could be tuned up a bit. Any thoughts on how to make improvements is greatly appreciated. First off,} there's a ton of image and word overlap on this story, which leads to a lot of writing and visibility issues. There definitely should have been done more body alignment, with the characters looking like more anthropomorphic animal animals more often. Additionally, this idea if poorly worded, as the phrase "spider" was spelled multiple times throughout the story and is such a terribly named word. It should be the appropriate sound, and become "f*cking spider"! It also doesn't help that you are writing in third person, which means that you could have brought in third person pronouns, such as "She" or "I" (though you may want to change some of the wording, such as "She"). The last thing I would change about this story is the different lengthy, regular sentences that end so often. One of the most common techniques used for lengthening stories is italicizing, wherein you substitute a word with an italicized sound. \\
\midrule
\textbf{Choice}: paragraph2 \\
\textbf{Justification}: Paragraph 1 has several grammatical flaws. It also begins to ramble in places. Paragraph 2 includes pertinent details and completes the main subject.\\ \\

\textbf{Choice}: paragraph1 \\
\textbf{Justification}: Both are truly terrible on every metric. Paragraph 2 is so problematic on fluency that Paragraph 1 wins despite using the non-word "amn\'t." As far as coherence and information goes, they are equally dreadful, and neither makes any sense whatsoever.\\ \\

\textbf{Choice}: paragraph2 \\
\textbf{Justification}: 1 deviates halfway through the prompt and starts talking about a different subject matter almost seemlessly. It almost makes sense if you don't read it very closely. \\
\bottomrule
\end{tabularx}
\end{center}
\vspace{-10pt}
\caption{WritingPrompts example 2.}
\vspace{-12pt}
\label{fig:writingprompts-example2}
\end{figure*}

\begin{figure*}[ht]
\begin{center}
\small
\begin{tabularx}{\textwidth}{X}
\toprule 
Repetitive generation from greedy search: \\
\midrule
\textbf{Aliens start abducting humans} and turning them into "space zombies" in the year 2077. The film is set in the year 2077, and follows a group of survivors who are forced to live in a space station. The film is directed by Ridley Scott, who also wrote the script. The film is set to be released in the UK on November 22. The film is the first in a planned trilogy of films based on the novel by Robert Aickman. The novel is set in the year 2077, and follows a group of survivors who are forced to live in a space station. The film is set to be released in the UK on November 22. The film is the first in a planned trilogy of films based on the novel by Robert Aickman. The novel is set in the year 2077, and follows a group of survivors who are forced to live in a space station. \\ \\

\textbf{The scientists have discovered something terrible .} .. The scientists have discovered something terrible... The scientists have discovered something terrible... The scientists have discovered something terrible... The scientists have discovered something terrible... The scientists have discovered something terrible... The scientists have discovered something terrible... The scientists have discovered something terrible... The scientists have discovered something terrible... The scientists have discovered something terrible... The scientists have discovered something terrible... The scientists have discovered something terrible... The scientists have discovered something terrible... The scientists have discovered something terrible... The scientists have discovered something terrible... The scientists have discovered something terrible... The scientists have discovered something terrible... The scientists have discovered something terrible... The scientists\\ \\

\textbf{The silence before the storm comes} from the people who are most affected by the storm. "The people who are most affected by the storm are the people who are most vulnerable to the storm," said Dr. David R. Jones, a professor of meteorology at the University of Alabama at Birmingham. "They are the people who are most likely to be hit by the storm." The storm is expected to bring heavy rain and strong winds to the Southeast, with the potential for flooding and mudslides. The National Weather Service said the storm is expected to bring up to a foot of rain to the Southeast, with the potential for up to a foot of rain in some areas. The storm is expected to bring up to a foot of rain to the Southeast, with the potential for up to a foot of rain in some areas. The storm is expected to bring up to a foot of rain to the Southeast, with the potential for up to a foot of\\
\midrule

Incoherent and nonsensical generations from unbiased sampling: \\
\midrule
\textbf{Aliens start abducting humans } but what begins as friendly relations turns sour as a vicious extraterrestrial force emerges with deadly power, and an army of killer cyborgs, sent back in time by the mysterious alien intelligence known as the Chronatrix. The first wave of an enemy race, that is coming for humanity. And only the best and bravest of them will survive the war. Written by James Hake\\ \\

\textbf{The scientists have discovered something terrible .} this is not a big deal. But the scientists are still in the stage to figure out what it is. It\'s not just a single thing. We\'re dealing with the biggest problem that mankind has facing it." So, what exactly does "this is not a big deal" mean? Well, there\'s no way to predict precisely what this "thing" will be. However, the physicists hope that the discovery will lead to some better method of analyzing supernovas, which could in turn lead to some advances in astronomy and technology. The discovery was made when an optical telescope belonging to the European Southern Observatory in Garching, Germany, was studying the aftermath of an exploding star. The supernova that caused the flare was so powerful that several billion people were nearby, according to the Associated Press, so the astronomers used a method previously used to observe supernovas from afar in telescopes orbiting Earth. "What we need now'\\ \\

\textbf{The silence before the storm comes} out of knots, flows easily, and, when it interferes, changes direction with an ebb and flow of its own accord. In place of ideal styles, this has come to be known as "Easy Song Coloration." Creative Music, and Learn to Run With it, presents Joshua Kelley Matheny, author of a new book inspired by classic cocktails. Only from CreativityAsGeek.com, drop in for some free advice on How to Make Bloody Marys, Cranberry Fish Tacos, Mud Pie Doughnuts, PSL s\'mores (all for less money than a bottle of Wild Turkey), and more. The first hundred requests can be monitored along with this week\'s theme between midnight and 12PM EST. Then a 100 laker may set sail for one of Prominent Icons at the Loop and Fillmore for a free party at the green fabric threshold. Author Joshua Kelley Matheny\'s'
\\
\bottomrule
\end{tabularx}
\end{center}
\vspace{-10pt}
\caption{Examples of \emph{de}generation. Text is generated by GPT2-large~\cite{radford2019language}.}
\vspace{-12pt}
\label{fig:problems}
\end{figure*}

\end{document}